\documentclass{article}

\PassOptionsToPackage{numbers, compress}{natbib}
\usepackage[final]{neurips_2021}

\usepackage[utf8]{inputenc} 
\usepackage[T1]{fontenc}    
\usepackage{hyperref}       
\usepackage{url}            
\usepackage{booktabs}       
\usepackage{amsfonts}       
\usepackage{nicefrac}       
\usepackage{microtype}      
\usepackage{xcolor}         

\usepackage{amsmath}
\usepackage{graphicx}
\usepackage{float}
\usepackage{tabularx}
\usepackage{subfig}
\usepackage{array}
\usepackage{wrapfig}

\title{Recurrent Off-policy Baselines for Memory-based Continuous Control}

\author{%
  Zhihan Yang\thanks{Equal contribution.} \\
  Carleton College \\
  Northfield, MN, USA \\
  \texttt{yangz2@carleton.edu} \\
   \And
   Hai Nguyen\footnotemark[1] \\
   Northeastern University \\
   Boston, MA, USA \\
   \texttt{nguyen.hai1@northeastern.edu} \\
}

\begin{document}

\maketitle

\begin{abstract}
When the environment is partially observable (PO), a deep reinforcement learning (RL) agent must learn a suitable temporal representation of the entire history in addition to a strategy to control. This problem is not novel, and there have been model-free and model-based algorithms proposed for this problem. However, inspired by recent success in model-free image-based RL, we noticed the absence of a model-free baseline for history-based RL that (1) uses full history and (2) incorporates recent advances in off-policy continuous control. Therefore, we implement recurrent versions of DDPG, TD3, and SAC (RDPG, RTD3, and RSAC) in this work, evaluate them on short-term and long-term PO domains, and investigate key design choices. Our experiments show that RDPG and RTD3 can surprisingly fail on some domains and that RSAC is the most reliable, reaching near-optimal performance on nearly all domains. However, one task that requires systematic exploration still proved to be difficult, even for RSAC. These results show that model-free RL can learn good temporal representation using only reward signals; the primary difficulty seems to be computational cost and exploration. To facilitate future research, we have made our PyTorch implementation publicly available\footnote{Code: \url{https://github.com/zhihanyang2022/off-policy-continuous-control}}.
\end{abstract}

\section{Introduction}

In recent years, deep off-policy reinforcement learning (RL) algorithms based on learning the optimal $Q$-function is enjoying great success in fully observable continuous control domains \cite{lillicrap2015continuous, fujimoto2018addressing, haarnoja2018soft, haarnoja2018soft2}. These algorithms usually achieve significantly better performance than deep on-policy RL methods given the same number of environment interactions and can learn highly competent policies on robotic locomotion tasks (e.g., MuJoCo \cite{todorov2012mujoco} domains in \cite{brockman2016openai}) where the state and action spaces are large. Their popularity
also comes from their easy-to-implement update rules derived from the Bellman equation, and their robustness to hyper-parameters.

In this paper, we attempt to extend their success to partially observable (PO) domains. Continuous control under partial observability is an interesting problem for two reasons. First, continuous action and partial observability are common in real-world applications and often occur together, e.g., in robotics. Second, PO domains are more challenging than fully observable domains because, to succeed, an RL agent must learn a suitable temporal representation of the entire observation-action history, in addition to a strategy to control.

\begin{figure}%
    \centering
    \includegraphics[width=0.75\textwidth]{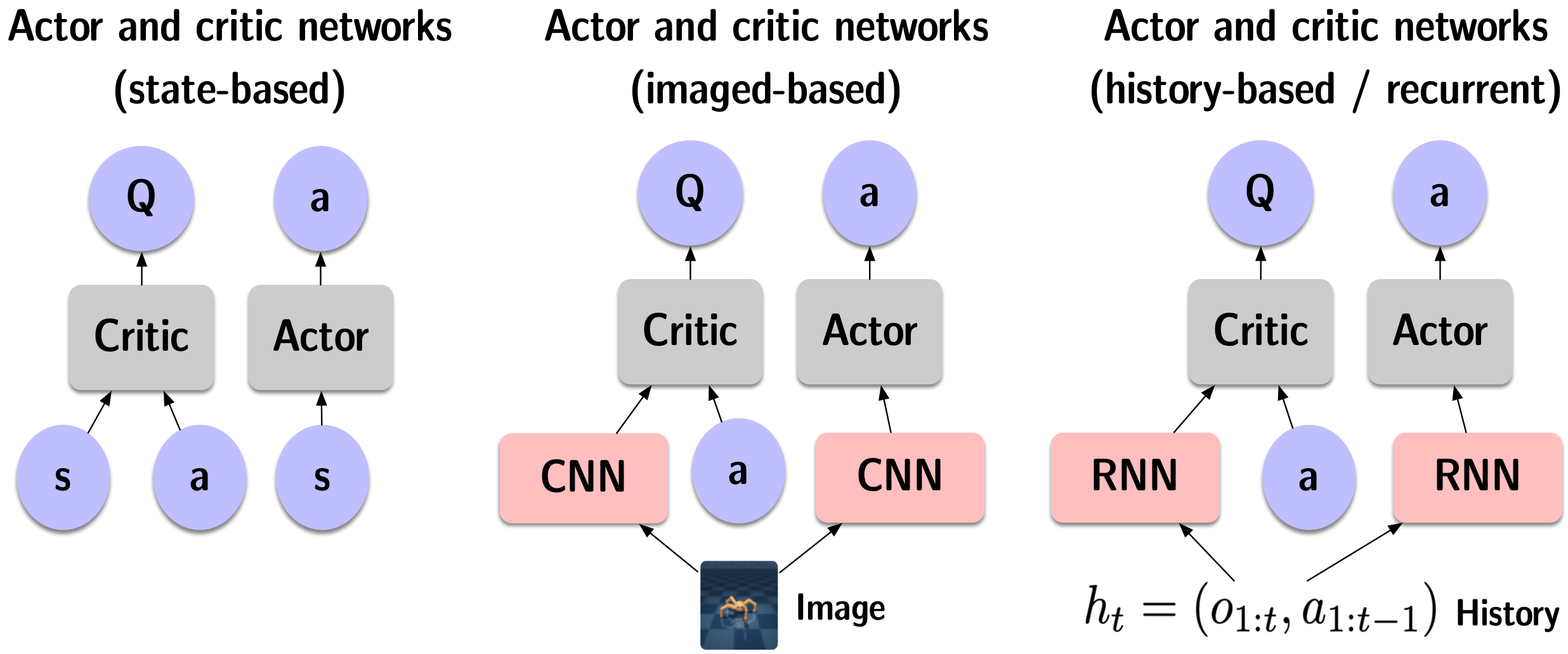}
    \caption{Comparison between architectures for standard, image-based, and recurrent off-policy model-free RL. The convolutional neural networks (CNNs) and recurrent neural networks (RNNs) can be trained end-to-end using gradients from loss terms, just like the actor and critic networks.}%
    \label{fig:intro_figure}%
    \vspace{-10pt}
\end{figure}

Continuous control under partial observability is not a novel problem, and there have been model-free and model-based algorithms proposed for this problem (see Section \ref{related-work}). However, we recognize the need for an open-source implementation of baseline algorithms that (1) are easy to implement, (2) can learn from the entire history, and (3) can learn stably and achieve good performance on a diverse set of PO tasks. To fill in the first and second gap, this paper describes a neural network architecture (see the rightmost of Figure \ref{fig:intro_figure}) that can be used to easily implement recurrent versions of DDPG, TD3, and SAC (RDPG, RTD3, and RSAC), and draws connection to a state-of-the-art image-based off-policy model-free algorithm DrQ \cite {kostrikov2020image} (see the middle of Figure \ref{fig:intro_figure}). This architecture allows the recurrent agents to take into account the entire history $h_t=\{o_{1:t},a_{1:t-1}\}$ at each timestep $t$, making it possible to learn arbitrary time dependencies. In addition, we adapt the standard experience replay procedure for these recurrent agents. To our knowledge, these agents are not available \textit{as is} in popular RL repositories \cite{stable-baselines, Raffin_Stable_Baselines3_2020, JMLR:v22:20-376, weng2021tianshou}.  

To fulfill the third gap, we evaluate RDPG, RTD3, and RSAC on short-term and long-term PO domains. In addition, we investigate important design choices, including the sharing of recurrent representation across actors and critics, and the types of recurrent neural networks used. Our experiments show that RDPG and RTD3 can surprisingly fail on some domains and that RSAC is the most reliable, often reaching near-optimal performance on nearly all domains. However, one task that requires systematic exploration still proved to be difficult, even for RSAC. These results show that model-free RL can learn good temporal representation using only reward signals; the primary difficulty seems to be computational cost and exploration. In the end, we suggest future research directions. Note that we do not intend to compare our algorithms to other algorithms but show that one of our algorithms can achieve near-optimal performance in several domains.

\section{Related work}\label{related-work}

There have been several recent works on learning memory-based policies in deep RL for continuous control. Recurrent Deterministic Policy Gradient (RDPG) \cite{heess2015memory} prepends recurrent layers to both the actor and critic networks of Deep Deterministic Policy Gradient (DDPG) \cite{lillicrap2015continuous}, and was able to solve a variety of simple PO domains, including sensor integration and memory tasks. However, since it is based on DDPG, it inevitably suffers from value over-estimation and premature policy convergence, two problems addressed later by algorithms such as Twin-delayed DDPG (TD3) \cite{fujimoto2018addressing} and Soft Actor-critic (SAC) \cite{haarnoja2018soft}. Moreover, it has not been publicly reproduced. To address value over-estimation and investigate high-dimensional tasks, \cite{meng2021memory} combines TD3 and LSTM \cite{hochreiter1997long} into TD3-LSTM. While TD3-LSTM shows good performance on high-dimensional sensor integration tasks, its network architecture is designed for POMDPs that are solvable with very short-term memory (e.g., 3-5 timesteps) and cannot be efficiently extended. A concurrent work \cite{ni2021recurrent} also combines TD3 and LSTM, but can be used with arbitrary memory length and is hence most similar to our work.

While the aforementioned works learn the temporal representation and the policy jointly, Variational Recurrent Model (VRM) \cite{han2019variational} was proposed to learn these two separately and showed better performance than its SAC-LSTM implementation (see Appendix \ref{rsac-vrm}). However, learning a separate model of the environment is a complex task, and the investigated domains are exclusively sensor integration tasks with either positional or velocity information and dense rewards. In this sense, VRM is related to several model-based algorithms proposed for image-based control tasks \cite{lee2019stochastic, hafner2019learning, hafner2019dream}, since each image only offers positional information and an optimal policy must integrate such information across frames. However, learning from images entails a different set of challenges and is beyond our scope.

While previous approaches all fall under off-policy RL, various other approaches have been proposed for memory-based control in general. Several approaches build on the concept of belief states \cite{kaelbling1998planning}, a distribution over the state space given the entire history. Since the precise belief state is computationally expensive to track for continuous domains, algorithms mostly either work with small and discrete observation and action spaces (e.g., \cite{nguyen2020belief}) or only seek to approximate the belief (e.g., with particle filters \cite{igl2018deep}, \cite{ma2020discriminative}). While both approaches require the use of RNNs, \cite{zhang2016learning} does not and instead lets the policy read and set part of the state space as a form of memory and train the policy via directed policy search. However, it uses a linear policy and requires a well-defined latent space during training. Apart from continuous control, memory-based control in discrete action spaces has been explored mainly by adding recurrent layers to the $Q$-network \cite{hausknecht2015deep, zhu2017improving, lample2017playing} in DQN \cite{mnih2015human}.

\section{Background}

\subsection{Partially observable Markov Decision Processes (POMDP)}

In real-world settings, an agent often can only have access to a reflection of the underlying state of the environment due to noise, occlusions, limited measurement resolutions, and so on. Under partial observability, a decision-making problem can be effectively modeled as a POMDP \citep{kaelbling1998planning}.
We can formally specify a finite-horizon POMDP by the tuple
$(\mathcal{S}, \mathcal{A}, \mathcal{T}, \mathcal{R}, \Omega, \mathcal{O}, H, \gamma)$, where $\mathcal{S}$ is the state space, $\mathcal{A}$ is the action space, $\mathcal{T}(s_{t+1} | s_t,a_t)$ is the transition probability,  $\mathcal{R}(s_t,a_t,s_{t+1})$ is the reward function, $\Omega$ is the observation space, $\mathcal{O} (o_t| s_{t+1}, a_t)$ is the observation probability, $H$ is the horizon, and $\gamma \in [0,1]$ is the discount factor. The goal is to learn a policy $\pi$ that maximizes the expected discounted return for a finite horizon $H$ defined as $\mathbf{E}_\pi[\sum_{t=0}^{H} \gamma^t r_t]$. 
To perform optimally in a POMDP, an agent must condition its policy on the entire history of observations and actions that it has seen so far. However, the size of the history space increases exponentially with the length of the history. Therefore, a recurrent neural network (RNN) is often used to summarize the history, and the policy is then condition on the RNN's fixed-size hidden states.

\subsection{Off-policy RL methods for fully-observable continuous control} \label{background-offpolicy-rl}

\paragraph{Deep deterministic policy gradient (DDPG)} DDPG \cite{lillicrap2015continuous} is an extension of the tabular Q-learning to (1) continuous state spaces and (2) continuous action spaces. To deal with (1), it uses a deep neural network $Q_\phi$ to represent the $Q$-function, commonly known as the \textit{critic}. It replaces the exact maximization problem in target computation (which is costly due to (2)) with approximation using an actor neural network $\mu_{\theta}$, commonly known as the \textit{actor}. During training, given a batch of $M$ transitions $\{(s_i, a_i, r_i, s'_i, d_i)\}_{i=1}^M$ uniformly sampled from a replay buffer, DDPG performs \begin{align*} 
\phi &\leftarrow \phi - \eta \nabla_\phi \frac{1}{M} \sum_{i=1}^M \left(\underbrace{Q_\phi(s_i, a_i)}_{\text{prediction}} - \underbrace{(r_i + \gamma (1 - d_i) Q_{\phi_{\text{targ}}}(s'_i, \mu_{\theta_{\text{targ}}}(s'_i)))}_{\text{target}}\right)^2 \tag{train critic} \\
\theta &\leftarrow \theta +\eta \nabla_{\theta} \frac{1}{M} \sum_{i=1}^{M} Q_{\phi}\left(s_i, \mu_{\theta}\left(s_i\right)\right) \tag{train actor},
\end{align*} 
where $\gamma$ is the discount factor, $\eta$ is the learning rate, and $\phi_{\text{targ}}$ and $\theta_{\text{targ}}$ are slowly changing versions of $\phi$ and $\theta$ obtained through polyak averaging.

\paragraph{Twin-delayed DDPG (TD3)} TD3 \cite{fujimoto2018addressing} addresses the issue of value over-estimation in DDPG. Specifically, in DDPG, $\mu_{\theta_{\text{targ}}}$ may exploit\footnote{This is because $\mu_\theta(s)$ is trained to maximize $Q_{\phi}(s, \cdot)$ and may exploit its erroneously high values; $\mu_{\theta_{\text{targ}}}(s)$ and $Q_{\phi_{\text{targ}}}(s, \cdot)$ are slowly changing versions of $\mu_\theta(s)$ and $Q_{\phi}(s, \cdot)$, and hence may inherit this property.} erroneously high values of $Q_{\phi_{\text{targ}}}(s, \cdot)$ and hence make the targets too large. As a result, bad actions at certain states may end up having over-estimated $Q$-values. In response, TD3 introduces three tricks. First, random noise is added to the output of $\mu_{\theta_{\text{targ}}}(s)$; this is called target smoothing. Second, $\mu_\theta$ is trained less frequently than $Q_\phi$; this is called delayed policy update. Finally, an additional $Q$-network is added, and both $Q$-networks are updated using the same target, computed using the minimum of the target versions of themselves; this is called clipped double-$Q$ learning. The actor is updated with respect to only one $Q$-network. TD3 significantly out-performs DDPG on several dense-reward high-dimensional continuous-control domains.

\paragraph{Soft Actor-Critic (SAC)} Unlike DDPG and TD3, SAC \cite{haarnoja2018soft} was derived from Soft Policy Iteration instead of the standard policy iteration. Nevertheless, SAC is similar to TD3 in two important ways. First, SAC learns a stochastic actor network, so its next-state actions used for target computation are sampled from a distribution, similar to target smoothing in TD3. Second, SAC also maintains two $Q$-networks, but uses both of them for training the actor network. However, SAC does not use target actor network or delayed policy update. In practice, TD3 and SAC have similar performance.

\subsection{Recurrent neural networks}

Recurrent neural networks (RNNs) can be best thought as a function that, at each timestep $t$, takes in (1) the previous summary ($\text{summary}_{t-1}$) and (2) some new information ($x_t$), and produces (1) some output ($\text{output}_t$) for subsequently layers and (2) some summary for the next timestep ($\text{summary}_{t}$). Despite this simple setup, several parametrizations have been proposed for learning this function effectively, with the most popular ones being Vanilla RNN (VRNN)  \cite{elman1990finding}, Long Short-Term Memory (LSTM) \cite{hochreiter1997long}, and Gated Recurrent Unit (GRU) \cite{cho2014learning}. For more details, please refer to Section \ref{appendix-rnn}. Since RNNs are differentiable with respect to their parameters, they can be trained using gradient descent in a procedure known as Backpropagation Through Time (BPTT).

\section{Architecture \& algorithm}

In this section, we describe the network architecture and algorithms that we used to train RDPG stably. We outline RDPG instead of RTD3 or RSAC because DDPG uses the least number of tricks for improving stability. Then, we describe how it can be easily extended for RTD3 and RSAC.

\paragraph{Architecture} DDPG involves the interaction between an actor $\mu_\theta(s)$ and a critic $Q_\phi(s, a)$. To adapt for POMDPs, recall that $h_t=(o_{1:t},a_{1:t-1})$ approximates $s_t$ in POMDPs. Therefore, we modify the actor to be $\mu_\theta(h_t)$ and the critic to be $Q_\phi(h_t,a_t)$. However, it is in general not possible to pass $h_t$ as is since fully-connected neural networks do not accept variable-sized inputs. To that end, we applied RNNs to distill and summarize $h_t$ into a fixed-size hidden vector $\hat{h}_t$. We also modified the standard experience replay to store individual episodes instead of individual transitions.

\paragraph{Algorithm} Formally, each update step consists of the following three sub-steps. First, one episode (a larger batch of episodes is used in practice) is sampled uniformly from the replay buffer:
\begin{align*}
\left(o_{1}, \ldots, o_{T+1}\right),\left(a_{0}, \ldots, a_{T}\right),\left(r_{1}, \ldots, r_{T}\right),\left(d_{1}, \ldots d_{T}\right),
\end{align*}
where $d_t$ represents the episode termination flag and $a_0$ is the first dummy action. 

Second, using this episode, the hidden vectors can be computed by unrolling an RNN over the sequence of observations and actions as follows ($\|$ denotes concatenation):
\begin{align*} \hat{h}_{1}^{\phi}, \ldots, \hat{h}_{T+1}^{\phi} &=\operatorname{RNN}_{\phi}\left(o_{1}\left\|a_{0}, \ldots, o_{T+1}\right\| a_{T}\right) \\ \hat{h}_{1}^{\phi_{\text {targ }}}, \ldots, \hat{h}_{T+1}^{\phi_{\text {targ }}} &=\operatorname{RNN}_{\phi_{\text {targ }}}\left(o_{1}|| a_{0}, \ldots, o_{T+1} \| a_{T}\right) \\ \hat{h}_{1}^{\theta}, \ldots, \hat{h}_{T+1}^{\theta} &=\operatorname{RNN}_{\theta}\left(o_{1}|| a_{0}, \ldots, o_{T+1}|| a_{T}\right) \\ \hat{h}_{1}^{\theta_{\text {targ }}}, \ldots, \hat{h}_{T+1}^{\theta_{\text {targ }}} &=\operatorname{RNN}_{\theta_{\text {targ }}}\left(o_{1}|| a_{0}, \ldots, o_{T+1} \| a_{T}\right), \end{align*}
where we used a separate RNN for each of the non-recurrent actor, the non-recurrent critic, and their target networks. 

Third, parameter updates to the recurrent actor and critic can be computed as follows:
\begin{align*}
\phi &\leftarrow \phi-\eta \nabla_{\phi}\left\{\frac{1}{T} \sum_{t=1}^{T}\left(Q_{\phi}\left(\hat{h}_{t}^{\phi}, a_{t}\right) - \left(r_{t}+\gamma\left(1-d_{t}\right) Q_{\phi_{\text {targ }}}\left(\hat{h}_{t+1}^{\phi_{\text {targ }}}, \mu_{\theta_{\text {targ }}}\left(\hat{h}_{t+1}^{\theta_{\text {targ }}}\right)\right)\right)\right)^{2}\right\} \\
\theta &\leftarrow \theta + \eta \nabla_{\theta} Q_{\phi}\left(\hat{h}_{t}^{\phi}, \mu_{\theta}\left(\hat{h}_{t}^{\theta}\right)\right)
\end{align*}
and $\phi_{\text{targ}}$ and $\theta_{\text{targ}}$ are updated using polyak averaging. Unlike in Section \ref{background-offpolicy-rl}, here $\phi$ denotes parameters of the \textit{entire} recurrent critic (the non-recurrent critic plus its corresponding RNN); the same logic applies to $\theta$, $\phi_{\text{targ}}$, and $\theta_{\text{targ}}$. Gradients flow through the hidden vectors and reach the RNN parameters, allowing the entire architecture to be trained end-to-end. For some environments, episodes have variable lengths, so masking is used.


\paragraph{Extending to RTD3 and RSAC} Unlike DDPG, TD3 and SAC use two critics instead of one. This difference is preserved in their recurrent versions, and hence they both use three RNNs (not counting the ones in target networks): one for the actor, two for the two critics. Tricks such as clipped double-$Q$ learning and entropy bonus rewards are directly ported from their non-recurrent versions.  

\section{Domains}

\begin{figure}
    \centering
    \setlength{\tabcolsep}{0pt}
    \subfloat[\texttt{pendulum}]{{\includegraphics[width=0.19\textwidth]{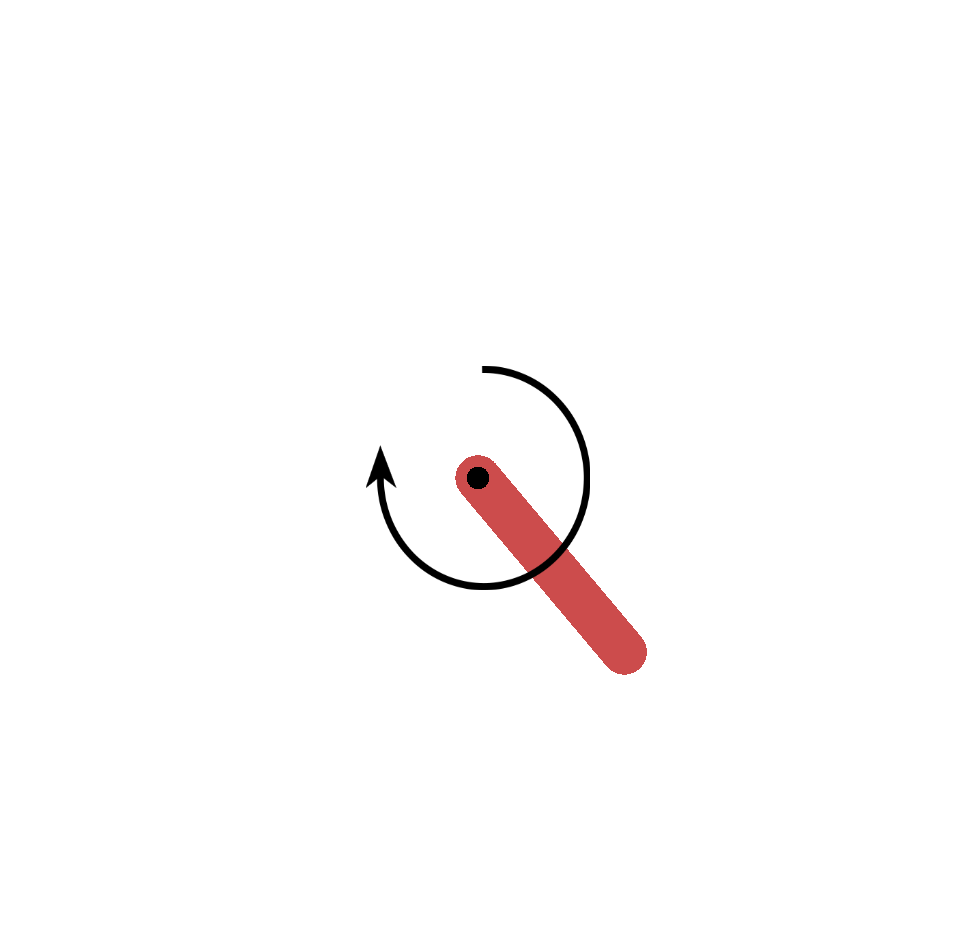} }}%
    \subfloat[\texttt{cartpole}]{{\includegraphics[width=0.19\textwidth]{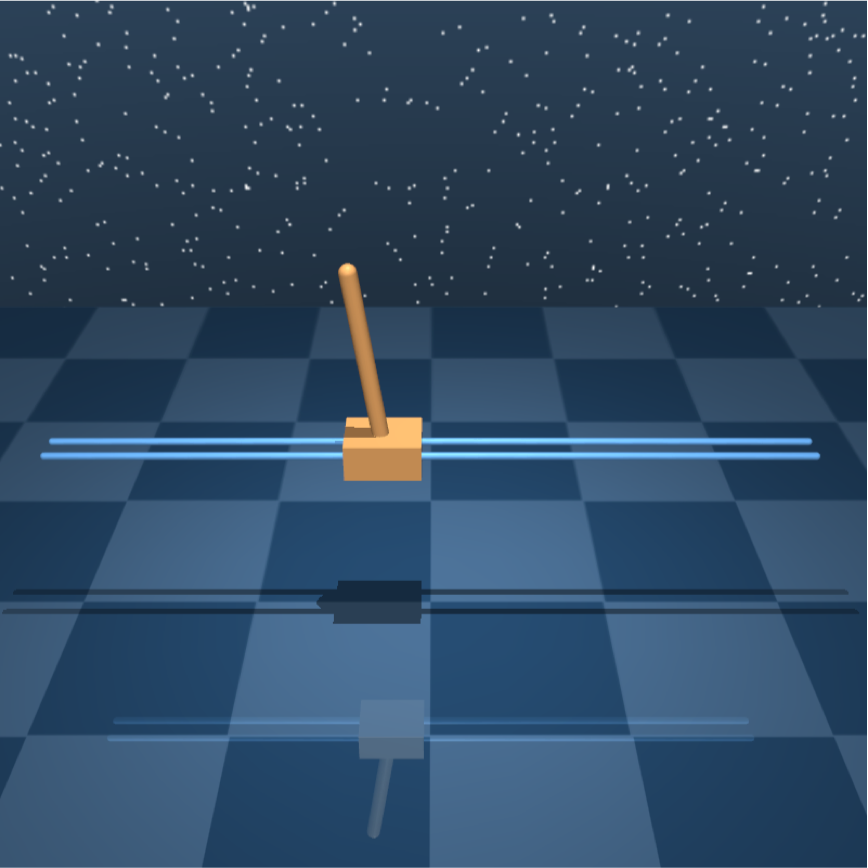} }}%
    \subfloat[\texttt{reacher}]{{\includegraphics[width=0.19\textwidth]{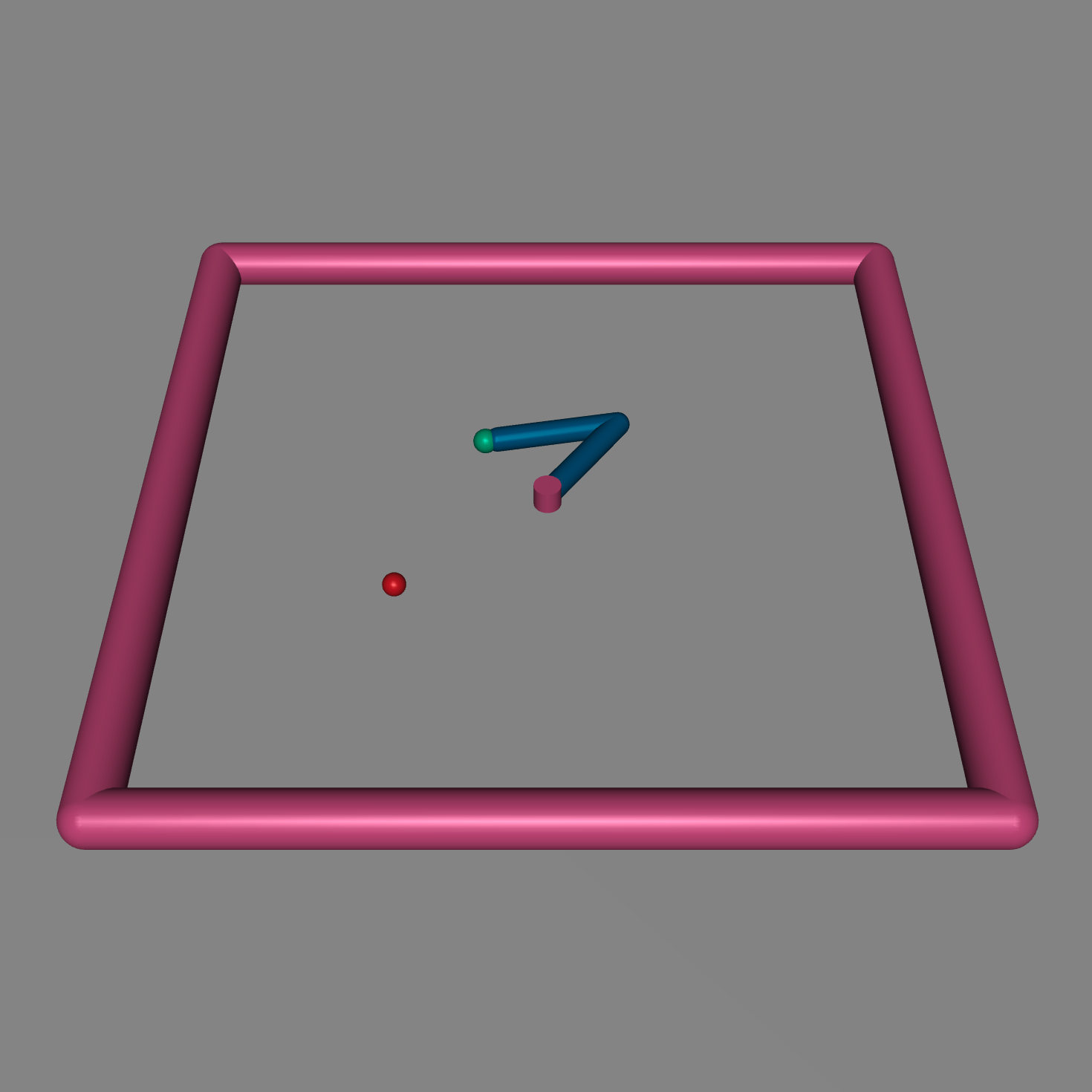} }}%
    \subfloat[\texttt{watermaze}]{{\includegraphics[width=0.19\textwidth]{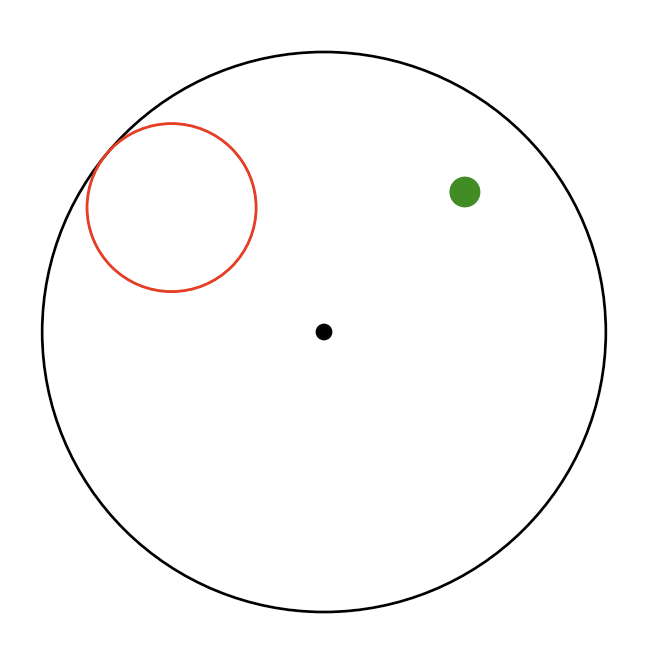} }}%
    \subfloat[\texttt{push-r-bump}]{{\includegraphics[width=0.19\textwidth]{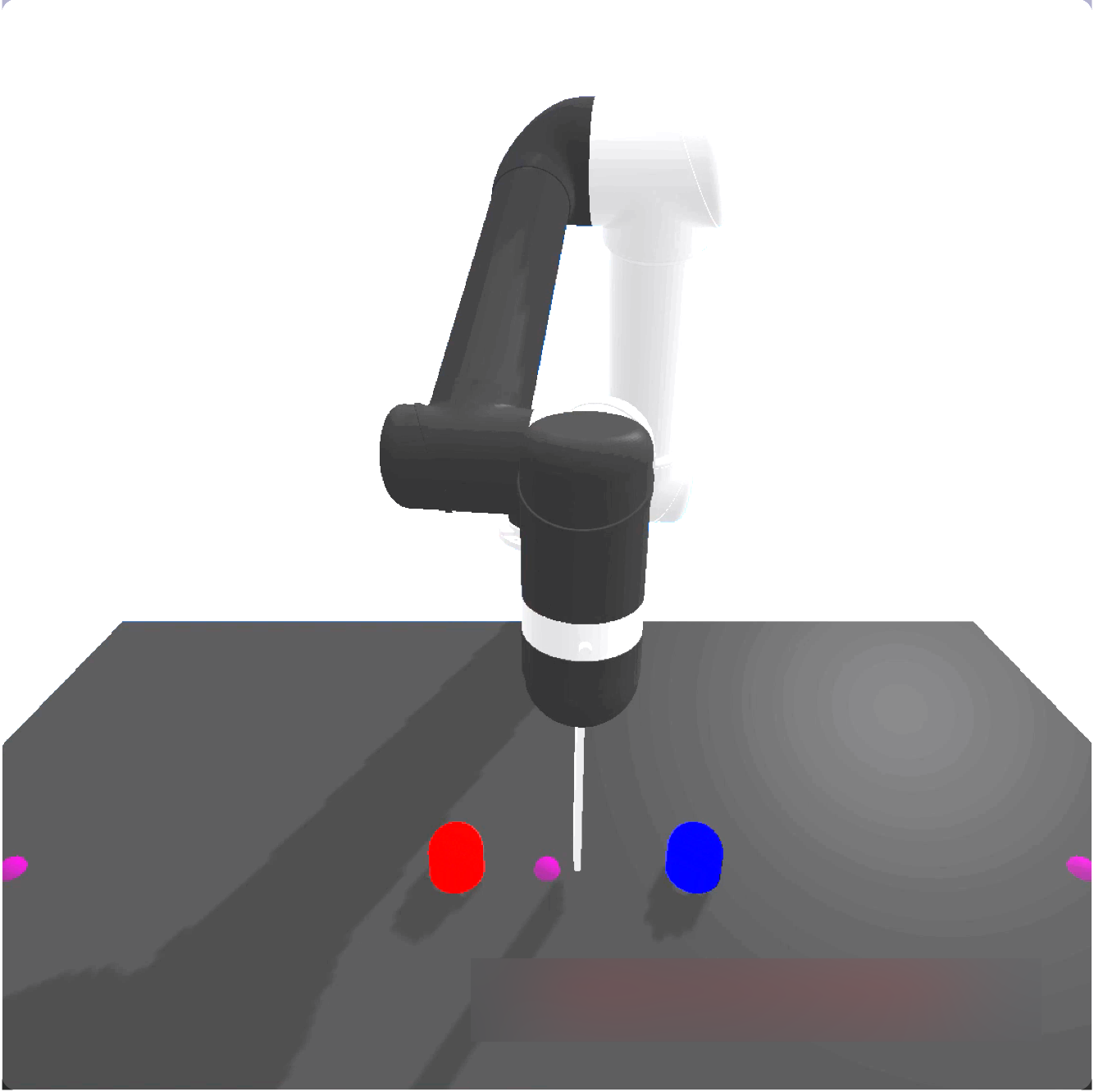} }}%
    \caption{Domains to perform experiments. For \texttt{watermaze} (d), the agent is illustrated as the green dot and the platform as the red circle.}%
    \label{fig:domains}%
\end{figure}

\subsection{Sensor integration tasks} 

Optimal control of physical systems might require integrating sensor inputs over a few past timesteps due to noise or the difficulty of measuring correctly all relevant physical quantities. Here, we use the position/velocity-only versions of three fully-observable physical-control domains (see a, b of Figure \ref{fig:domains}), similar to domains investigated in \cite{heess2015memory}, \cite{han2019variational}, and \cite{meng2021memory}. These domains are \texttt{pendulum-swingup} ($\dim(\mathcal{S})=3, \dim(\mathcal{A})=1$), \texttt{cartpole-balance} ($\dim(\mathcal{S})=5, \dim(\mathcal{A})=1$), and \texttt{cartpole-swingup} ($\dim(\mathcal{S})=5, \dim(\mathcal{A})=1$); they are generally considered to be increasingly more difficult. They all have a maximum episode length $T$ of $200$ and use a dense reward function. Their position-only versions remove all entries of the observation related to velocity, and their velocity-only versions remove all entries related to position. We also append the previous action to the velocity-only observations, which we observed to improve the performance of recurrent agents.

\subsection{Memory tasks}

Another important type of PO tasks are memory tasks, where the agent needs to memorize (instead of simply integrate) some information over timesteps. We evaluate our agents on two memory tasks: \texttt{reacher-pomdp} and \texttt{watermaze}. \texttt{reacher-pomdp} (see Figure \ref{fig:domains}c) ($T=50$) modifies the original 2-joint reacher task from OpenAI Gym \cite{brockman2016openai} so that the location of the goal only appears for the first timestep. This task provides dense rewards. \texttt{watermaze} (see Figure \ref{fig:domains}d) ($T=200$) is borrowed from \cite{heess2015memory}. Since the paper does not open-source its domains, we re-implemented it with minor modifications. \texttt{watermaze} simulates a task that is used to study spatial learning for rodents. In this task, the agent (e.g., a mouse) is released at the center of a circular world. It is rewarded by being on top of an invisible circular platform, which is randomly positioned per episode. The agent can only observe the platform when being on top of the platform, and is taken back to the center afterwards. To maximize return, the agent must memorize the location of the platform to return to it faster.

\subsection{Active exploration task}\label{active_explore_task}
\begin{figure}[H]
    \centering
    \includegraphics[width=0.75\textwidth]{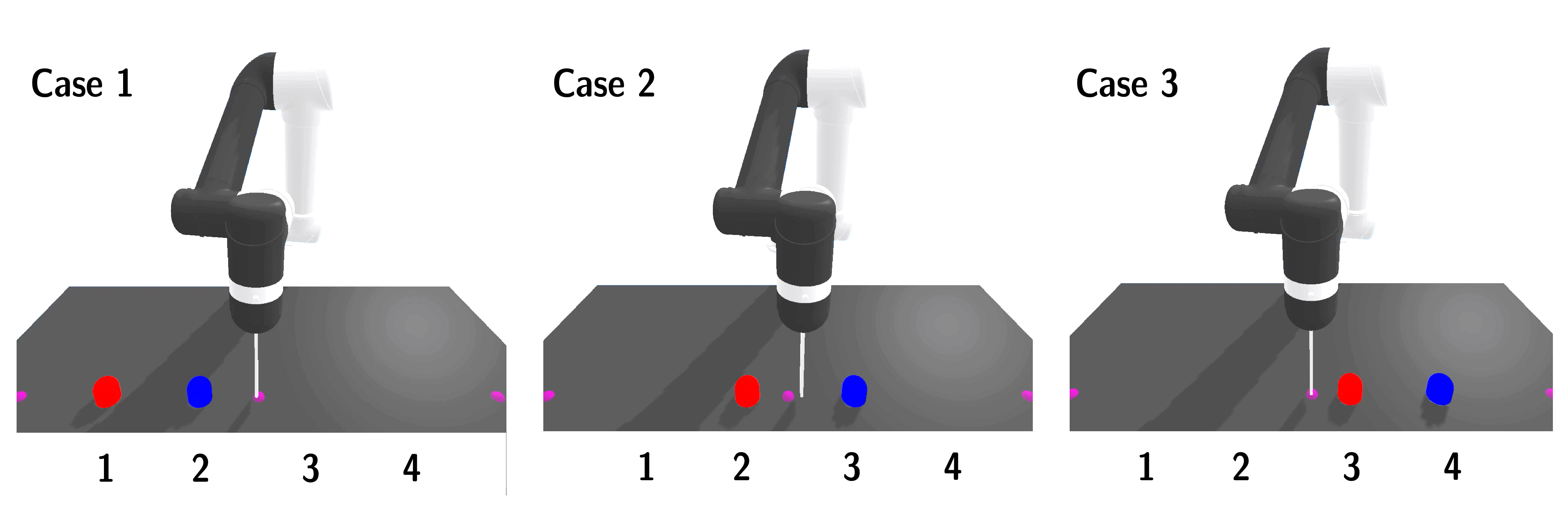} 
    \caption{Three cases in \texttt{push-r-bump}; colors and numbers are for illustration and are not observed.}%
    \label{fig:push_r_bump_cases}%
    \vspace{-10pt}
\end{figure}

In a PO world, it is very common for an agent to perform non-rewarding but informative actions to be more certain about the world before performing rewarding actions. As an example, we created a domain called \texttt{push-r-bump} (see Figure \ref{fig:domains}e) ($T=50$), which is a continuous version of a similar domain in \cite{nguyen2020belief}. In this domain, a robot controls the movement and the stiffness of a finger. When the finger is soft, it can glide over bumps without moving them, causing reactive angles on the finger. In contrast, when the finger is stiff enough, it can move bumps. Any bump movement will terminate the current episode and will bring a reward of $1$ if the right bump is pushed right for a short distance and $-1$ otherwise. The robot can observe the finger's position and angle, but it does not observe the bumps' locations. Starting each episode, the position of the finger is initialized around the center, and the two bumps are randomized to one of three possible cases (see Figure \ref{fig:push_r_bump_cases}). Therefore, an optimal agent must first use a soft finger to scan around and localize the right bump, and finally push this bump to the right for a short distance with a stiff finger to get a positive reward. 

\section{Experiments \& results}\label{exp-res}

Unless otherwise noted, (1) RDPG, RTD3, and RSAC take in observations instead of states, and (2) the default RNN used is LSTM. For return, we plot the mean and min/max (1 standard deviation for other statistics) w.r.t. 5 seeds as in \cite{haarnoja2018soft}. Statistics are recorded once per 1k steps (1k environment steps and 1k update steps). Each return recorded is the average return over 10 episodes with deterministic actors. Training time is w.r.t. Nvidia's RTX 2080 (16G). Hyper-parameters are in Appendix \ref{appendix-hyperparameter}.

\subsection{Performance on sensor integration tasks}

From the 3rd and 4th column of Table \ref{table:sensor_int}, we see that RSAC is the only algorithm to achieve similar performance at convergence as \texttt{SAC-state} on all PO domains. Interestingly, RDPG has 2 failed seeds in \texttt{pendulum-p} and failed completely in \texttt{cartpole-swingup-p/v}.  It is perhaps surprising that RDPG can have failed seeds on \texttt{pendulum-p}, because the MDP version of \texttt{pendulum-p} is arguably one of the simplest continuous control domains. 

It is well know that, in DDPG, the actor can exploit erroneously high values in the critics and destabilize learning \cite{fujimoto2018addressing}. Since the actor in RDPG is augmented with a two-layer RNN, the actor is more expressive and may hence be a stronger exploiter of such errors. However, since RDPG can perform reasonably well (despite one failed seed in \texttt{cartpole-swingup}) on the MDP versions of the domains (see the second column of Table \ref{table:sensor_int}), we argue that this source of exploitation, if exists, is not the primary cause for the algorithm to diverge on PO variants. Instead, we hypothesize that partial observability itself may open up more room for such exploitation, and is worthy of further investigation. A supporting evidence is that RTD3 and RSAC have successful seeds on \texttt{cartpole-swingup-p/v}, presumably because of their mechanisms designed to address such exploitation. 

RTD3 performs better than RDPG, and has similar performance at convergence as RSAC on \texttt{pendulum-p/v} and \texttt{cartpole-balance-p/v}. However, RTD3 learns slower than RSAC on \texttt{pendulum-p/v}, and has one failed seed in both \texttt{cartpole-swingup-p/v}. Interestingly, \texttt{RTD3-state} also has two failed seeds in the MDP version of \texttt{cartpole-swingup}, suggesting that this failing behavior may be due to RTD3's brittleness with more expressive actor and critics instead of partial observability. Empirically, through visualizing the learned policies in \texttt{cartpole-swingup-p/v}, we noticed that RTD3 can get stuck in a suboptimal policy of quickly moving the cart to one side to briefly drag the pole to its horizontal position (away from the downward vertical position) to gain some reward. 

As aforementioned, RSAC is the only algorithm that matches its non-recurrent MDP performance on all PO tasks. Algorithmically, RSAC is highly similar to RTD3 in terms of many design choices, except that its actor maximizes its entropy in addition to the return. Entropy maximization can prevent premature convergence of the actor, and has been shown to explore equally rewarding avenues of behavior much better than simply adding random noise to a deterministic policy \cite{haarnoja2017reinforcement}. In fact, exploration may be crucial to learning an expressive temporal representation from scratch since better exploration leads to a wider distribution of observations and rewards, which prevents over-fitting due to limited experience. In practice, this may help RSAC to be less affected by a large network and learn more consistently than RTD3 (e.g., by escaping local optima).

\begin{table}
    \centering
    \setlength{\tabcolsep}{0pt}
    \begin{tabular}{cccc}
    
        \includegraphics[width=0.25\linewidth]{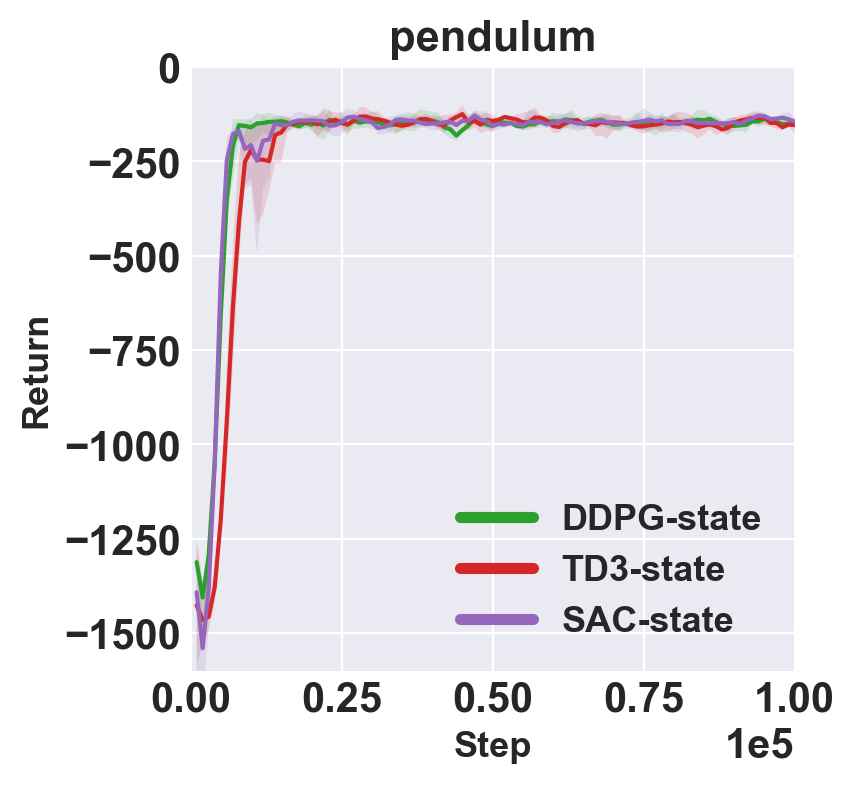}
        & \includegraphics[width=0.25\linewidth]{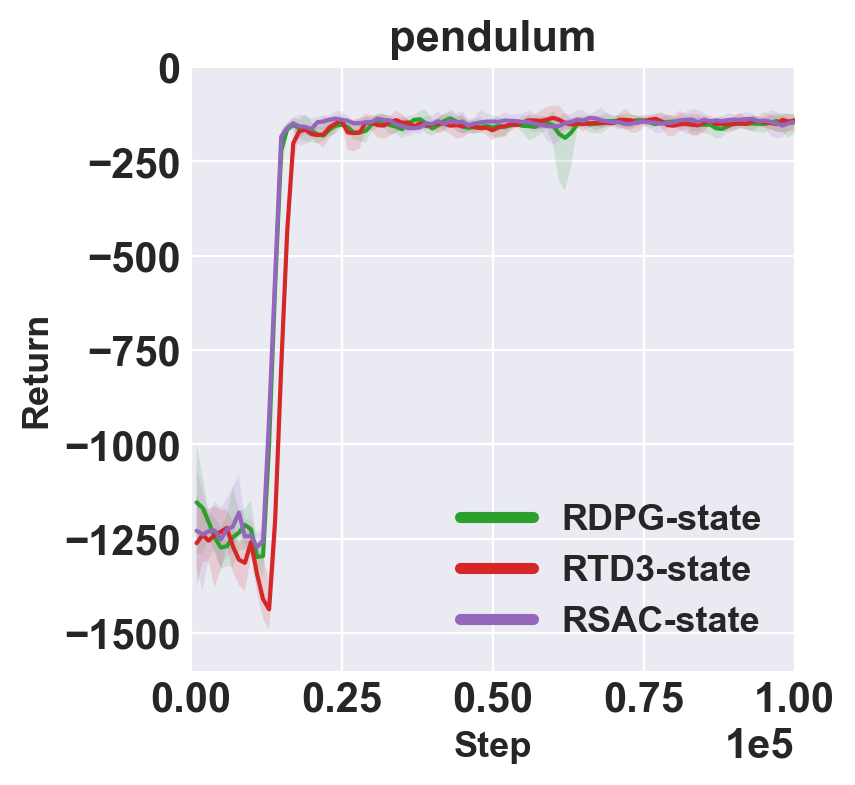}
        & \includegraphics[width=0.25\linewidth]{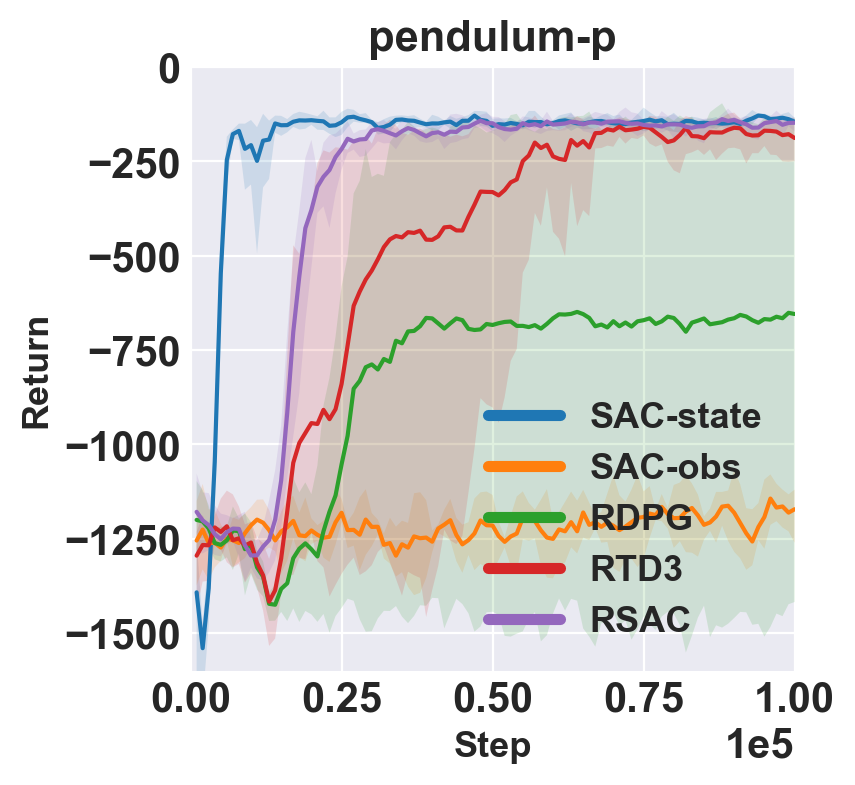}
        & \includegraphics[width=0.25\linewidth]{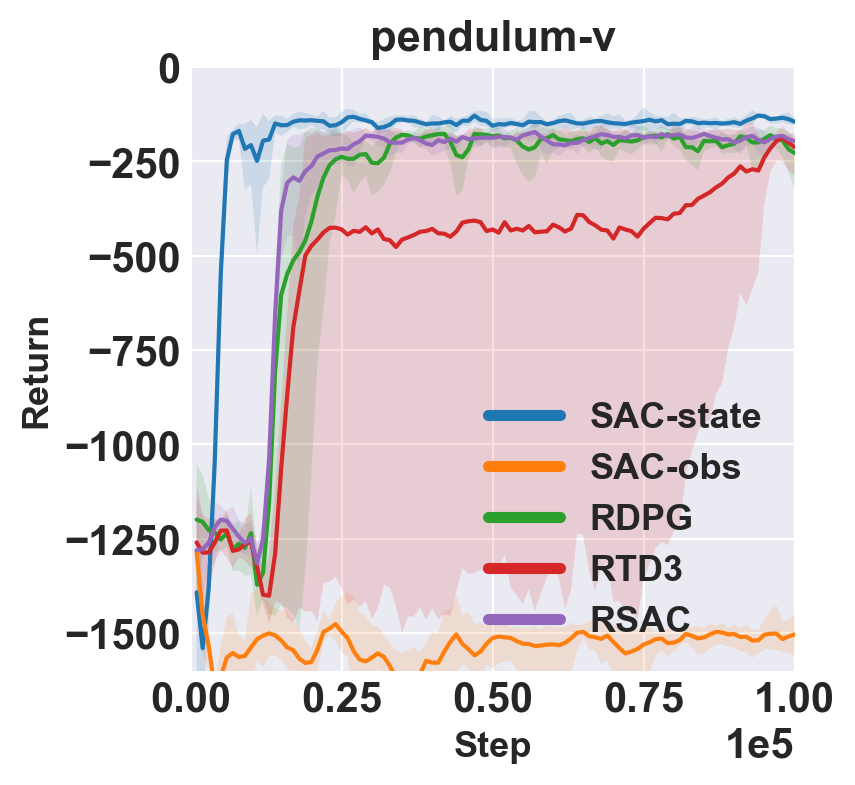}\\[-4pt]
        
        \includegraphics[width=0.25\linewidth]{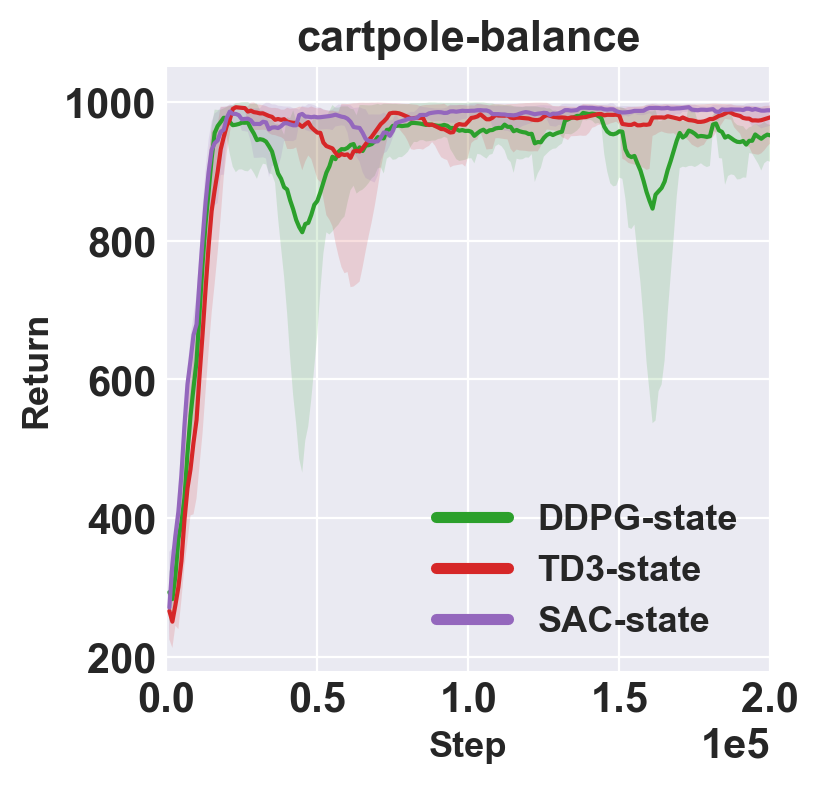}
        & \includegraphics[width=0.25\linewidth]{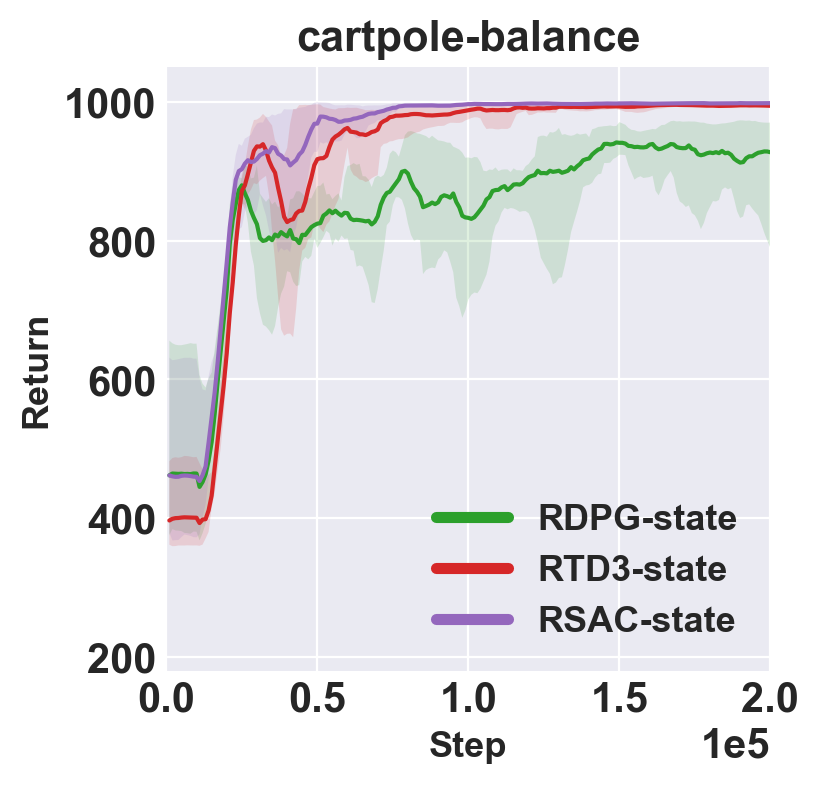}
        & \includegraphics[width=0.25\linewidth]{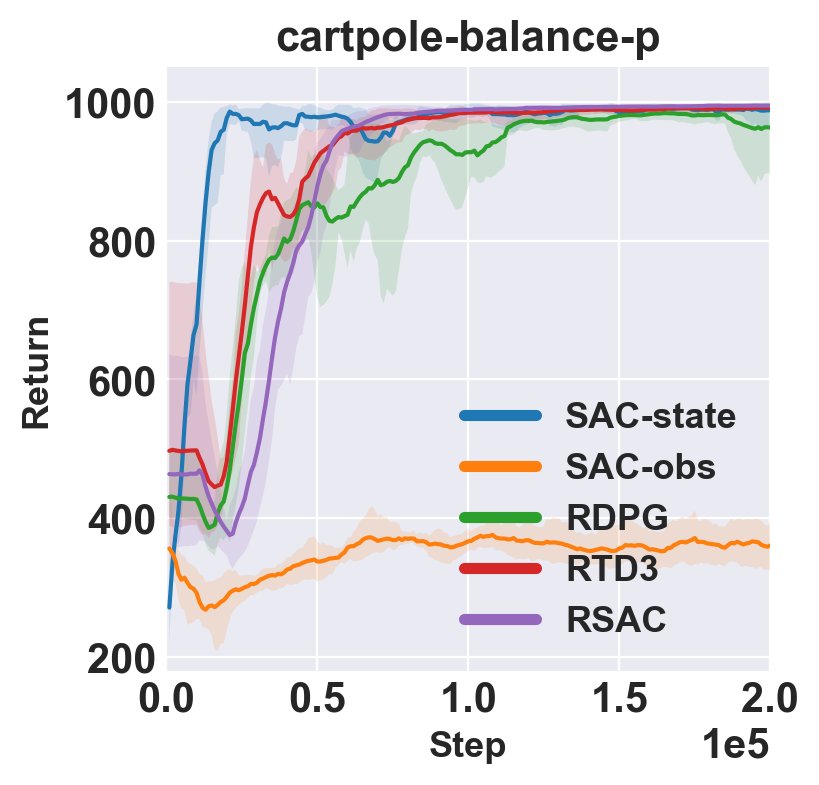}
        & \includegraphics[width=0.25\linewidth]{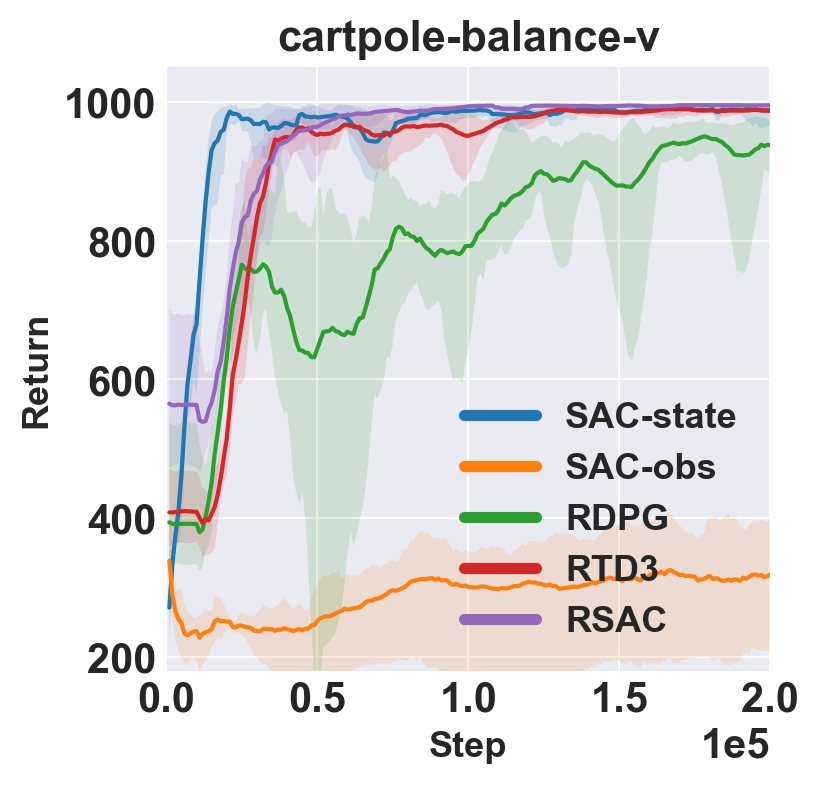}\\[-4pt]
        
        \includegraphics[width=0.25\linewidth]{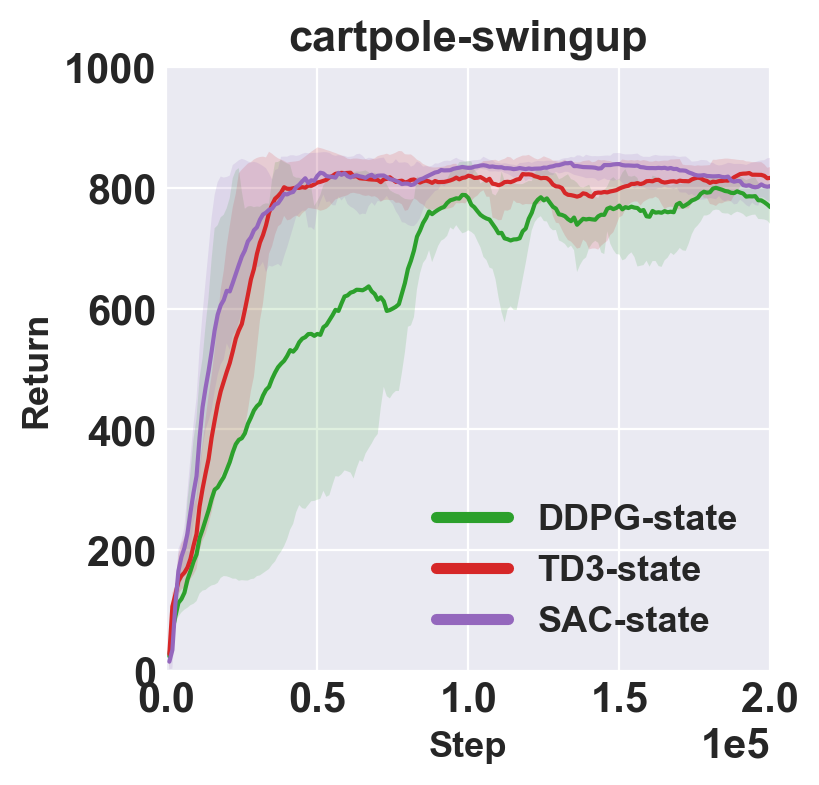}
        & \includegraphics[width=0.25\linewidth]{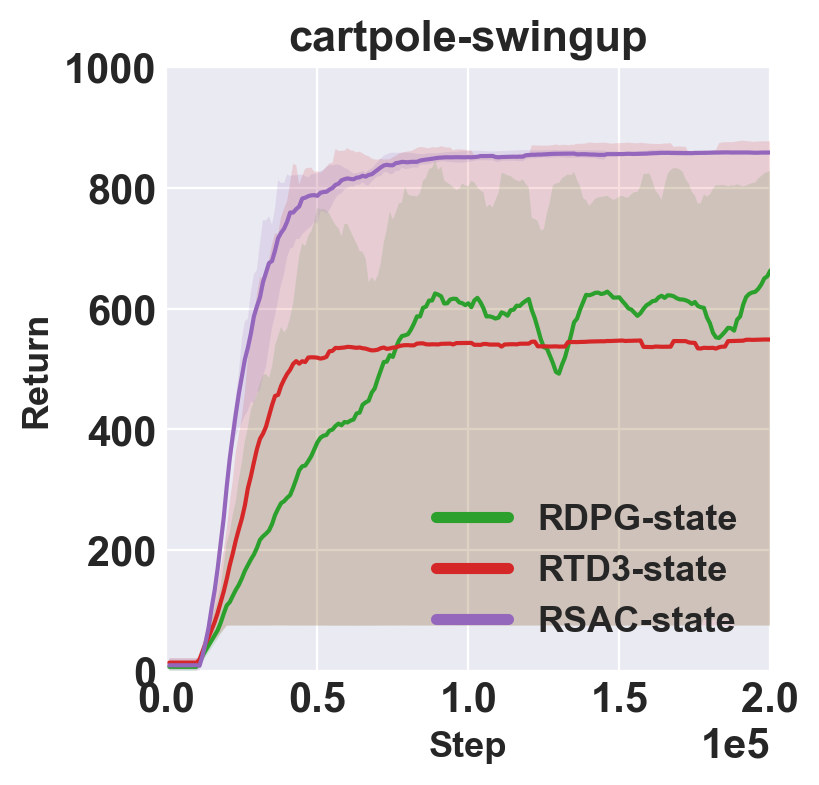}
        & \includegraphics[width=0.25\linewidth]{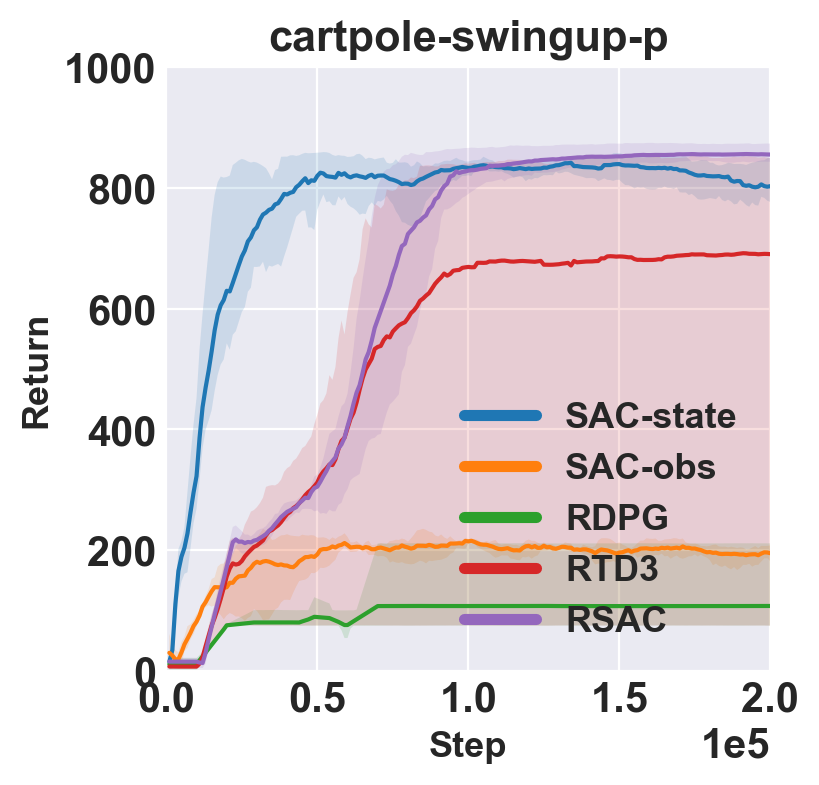}
        & \includegraphics[width=0.25\linewidth]{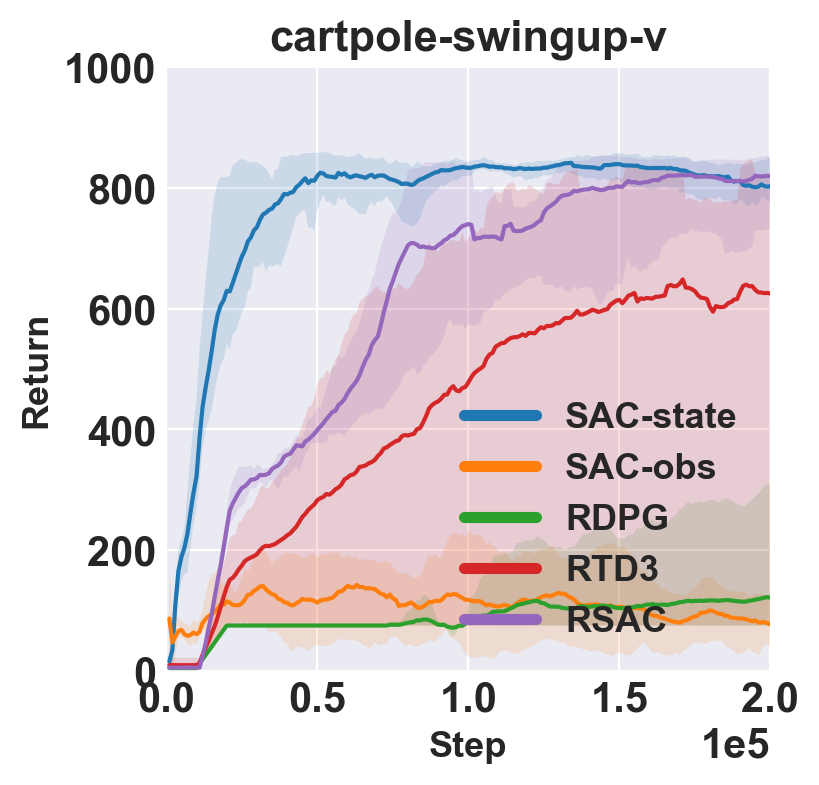}
        
    \end{tabular}%
    \caption{Performance on sensor integration tasks. DDPG, TD3, and SAC all learn competent policies given the state. However, the performance of DDPG and TD3 degrade with LSTMs - wide error bands are due to splits between successful and failed seeds.}
    \label{table:sensor_int}
    \vspace{-10pt}
\end{table}

\subsection{Performance and exemplary trajectories on memory and active exploration tasks}

\begin{figure}
    \centering
    \subfloat[Platform at $90^{\circ}$]{{\includegraphics[width=0.19\textwidth]{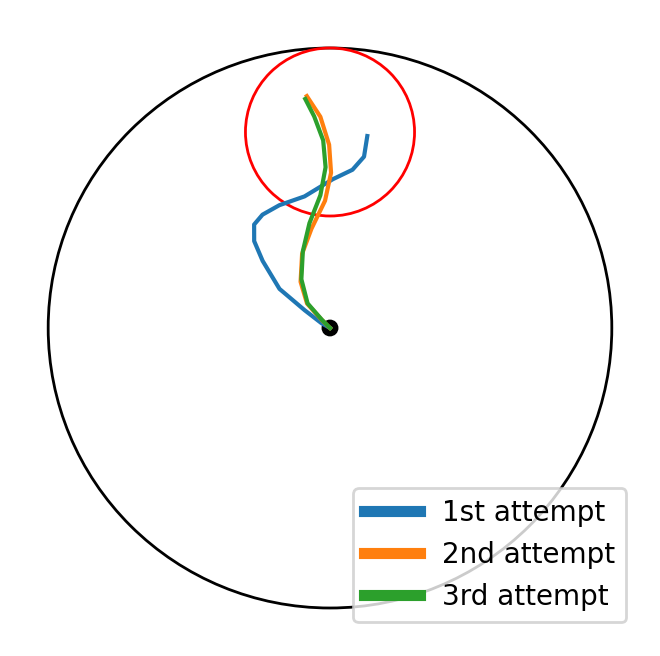} }}%
    \subfloat[Platform at $180^{\circ}$]{{\includegraphics[width=0.19\textwidth]{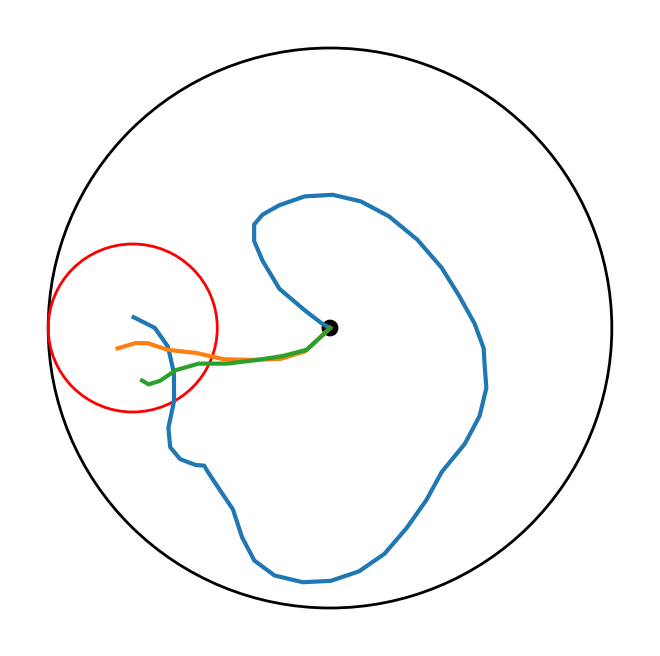} }}%
    \subfloat[Case 1 in Figure \ref{fig:push_r_bump_cases}; events happen from left to right.]{{\includegraphics[width=0.60\textwidth]{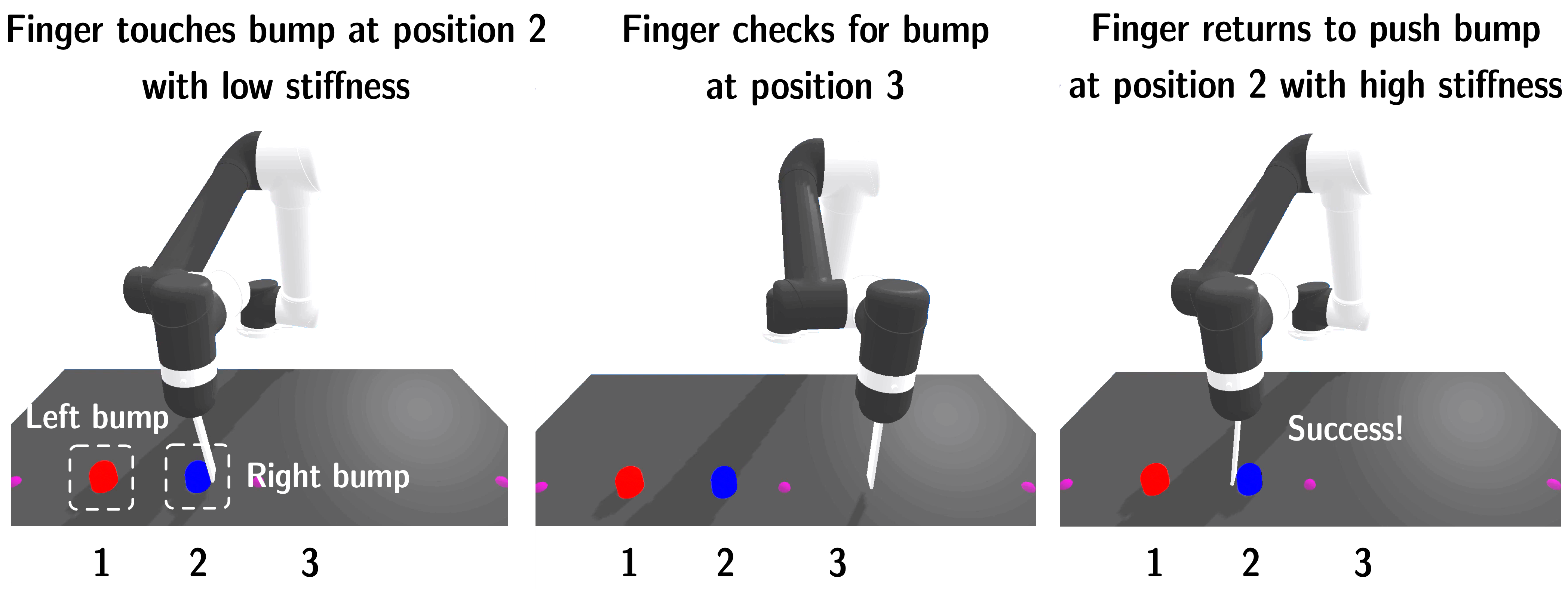} }}%
    \caption{Visualization of learned policies. (a) (b) One RSAC seed trained on \texttt{watermaze}. (c) One RTD3 seed trained on \texttt{push-r-bump} executing in one among three cases.}%
    \label{fig:policy_viz}%
    \vspace{-10pt}
\end{figure}

RDPG, RTD3, and RSAC all learn competent policies on \texttt{reacher-pomdp} (see Table \ref{table:memory_task_perf}), where dense rewards are available. In contrast, RSAC is the only algorithm that performs well and consistently on \texttt{watermaze}, which offers only sparse rewards and requires active exploration at the beginning of training. RSAC is trained under the entropy maximization framework, potentially making the policy explore better throughout training. We illustrate one RSAC seed in Figure \ref{fig:policy_viz}a and \ref{fig:policy_viz}b; it learns an exploratory policy and can memorize the platform position to return quickly within fewer timesteps.

In \texttt{push-r-bump}, no algorithm can consistently solve the task to 100\% success rate. Still, RDPG, RTD3, and RSAC all have successful seeds, and RTD3 and RSAC have seeds that solve 1 or 2 cases (among the 3 mentioned in Section \ref{active_explore_task}). This is likely an issue of insufficient exploration, since systematic exploration is required for the algorithms to solve all 3 cases correctly by chance to obtain sparse positive rewards to start learning. We visualize 1 successful RTD3 seed on 1 case in Figure \ref{fig:policy_viz}c, and the same seed on 2 other cases in Appendix \ref{appendix-traj} along with how RNN cell states evolve.

\begin{table}
    \centering
    \setlength{\tabcolsep}{0pt}
    \begin{tabular}{ccccc}
        \includegraphics[width=0.24\linewidth]{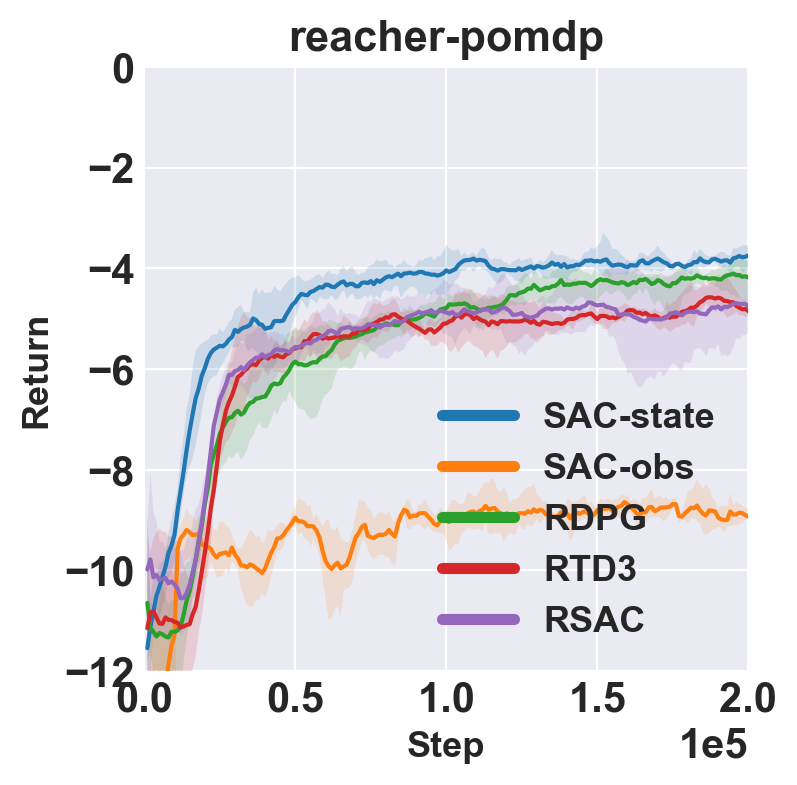}
        & \includegraphics[width=0.24\linewidth]{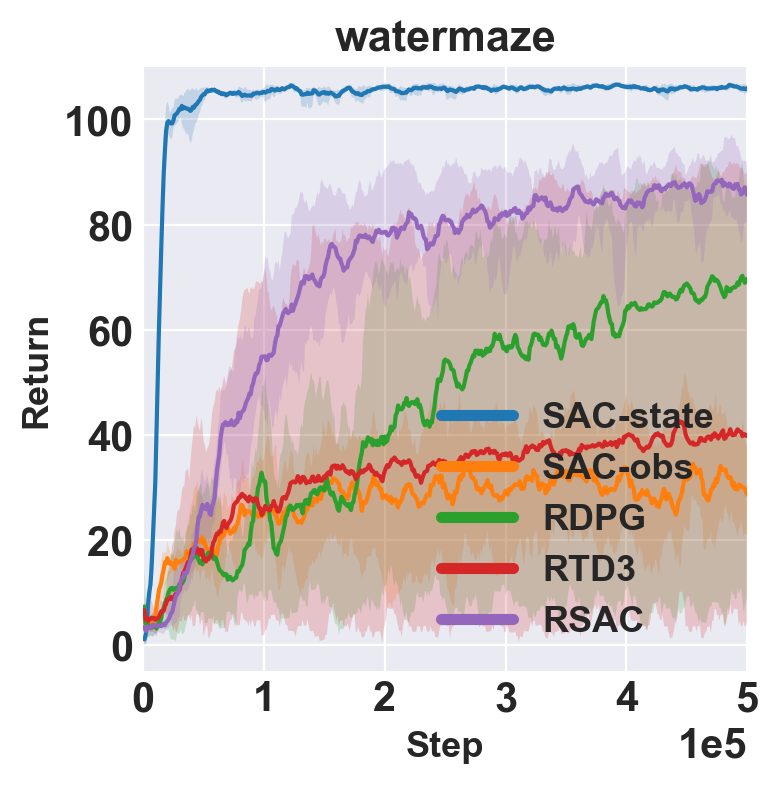}
        & \includegraphics[width=0.175\linewidth]{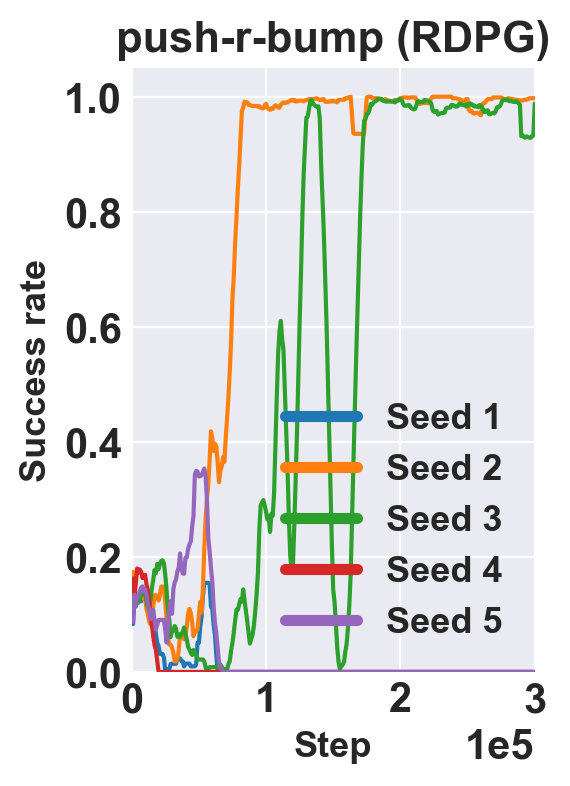}
        & \includegraphics[width=0.175\linewidth]{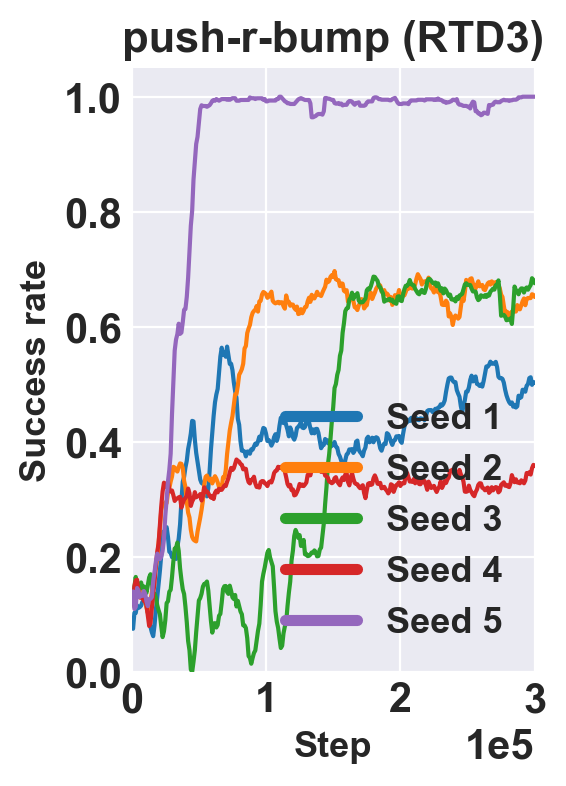}
        & \includegraphics[width=0.175\linewidth]{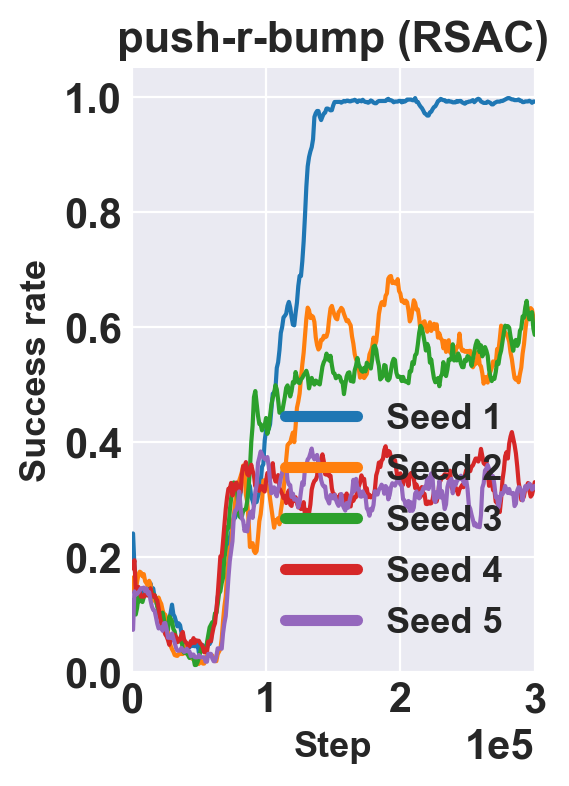}
    \end{tabular}%
    \caption{Performance on memory and active exploration tasks. For \texttt{push-r-bump}, we plot individual seeds' success rates to show that neither RDPG, RTD3, nor RSAC can consistently solve the task.}
    \label{table:memory_task_perf}
    \vspace{-10pt}
\end{table}


\subsection{Sharing a single recurrent representation across the actor and the critics} \label{section-share}

\begin{wrapfigure}{r}{0.25\textwidth}
  \begin{center}
    \includegraphics[width=0.25\textwidth]{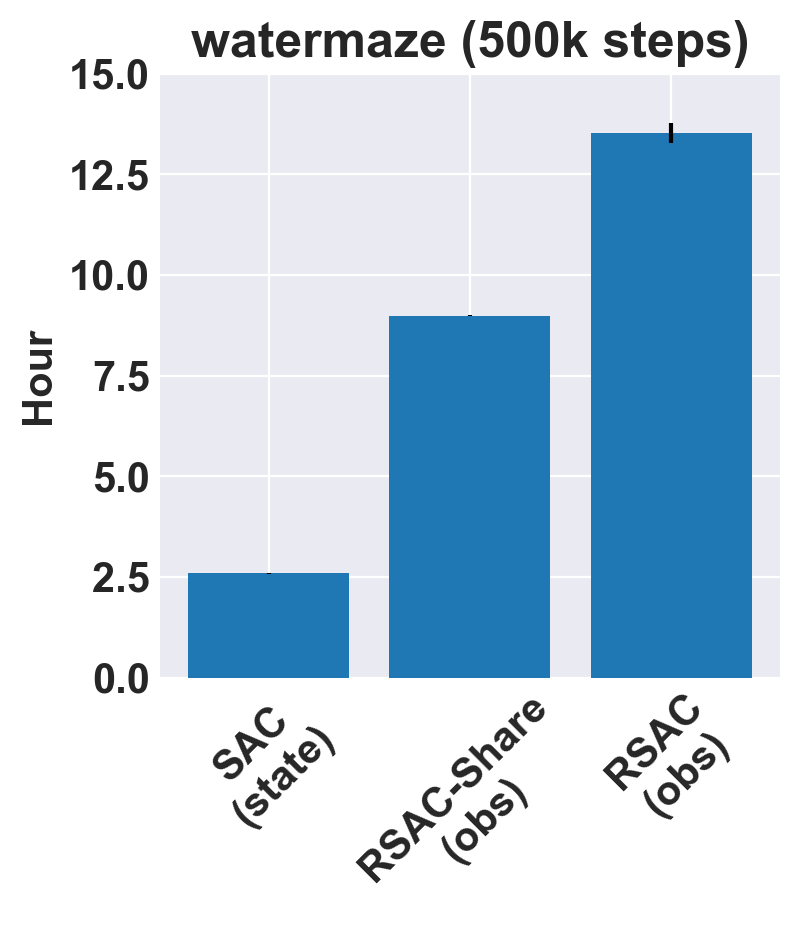}
  \end{center}
  \caption{RSAC-Share reduces training time.}
  \label{fig:share_time}
  \vspace{-10pt}
\end{wrapfigure}

In RSAC, there are 5 recurrent modules: 1 within the actor, 2 within the two critics, and 2 within the two critic targets. In this setup, each update step involves 5 rounds of RNN forward passes and 3 rounds of BPTTs. Since each round of computation scales with $T$, it is natural to ask whether sharing a \textit{single} recurrent module across them (1 for all non-target networks, and 1 for all target networks) could have similar performance as without sharing while using less computation. We call this approach RSAC-Share. While it is straightforward to optimize parameters of the single shared representation w.r.t. both the actor and the two critic losses, these two kinds of losses might have different scales and could lead to instability \cite{heess2015memory}. Additionally, while this approach only involves 2 rounds of RNN forward passes, it still requires 3 rounds of BPTTs.

Seeking further improvement, we borrow the insights from DrQ \cite{kostrikov2020image}, which trains SAC on image-based domains and shares a single convolutional neural network (CNN) across an actor and two critics. In DrQ, the shared CNN accumulates gradients from both critic losses; the gradients of the actor loss w.r.t. the CNN's parameters are \textit{ignored}. Similarly, for RSAC-Share, we optimize the parameters of the shared recurrent module w.r.t. the two critic losses only. Intuitively, using gradients from the critic losses (instead of the actor loss) helps the shared recurrent module learn a richer representation, since, given a state, the critics must learn values for all actions but the actor only needs to learn the maximizing action. Doing so also reduces computation: the final version of RSAC-Share uses 2 rounds of RNN forward passes and BPTTs per update step, halving the RNN computation in RSAC.

Our experiments show that RSAC-Share is on par with RSAC (see the first row in Table \ref{table:share}), and significantly reduces training time (see Figure \ref{fig:share_time}). RSAC-Share is less stable on \texttt{cartpole-swingup-v}, but reaches better performance at convergence on \texttt{reacher-pomdp}. 

We have two interesting observations. Firstly, the Pearson correlation (see Appendix \ref{pearson}) of $Q$-values produced by the two critics is higher for RSAC-Share (see the second row in Table \ref{table:share}), presumably because they share the same recurrent representation. Such high correlation potentially makes clipped double $Q$-learning in RSAC-Share ineffective, since the minimum of two highly correlated random variables is almost equal to either one of the two. Secondly, RSAC-Share learns higher $Q$-values than RSAC does (see the third row in Table \ref{table:share}); this is likely due to the first observation, and that the actor in RSAC-Share may be a stronger maximizer of the critics' outputs, since it directly acts on the critics' recurrent representation. While these observations suggest risks of value over-estimation, RSAC-Share has strong empirical performance, and out-performs both RDPG and RTD3. Note that, for both observations, $Q$-values are calculated from 10 episodes sampled from the replay buffer.

\begin{table}
    \centering
    \setlength{\tabcolsep}{0pt}
    \begin{tabular}{cccc}
        \includegraphics[width=0.25\linewidth]{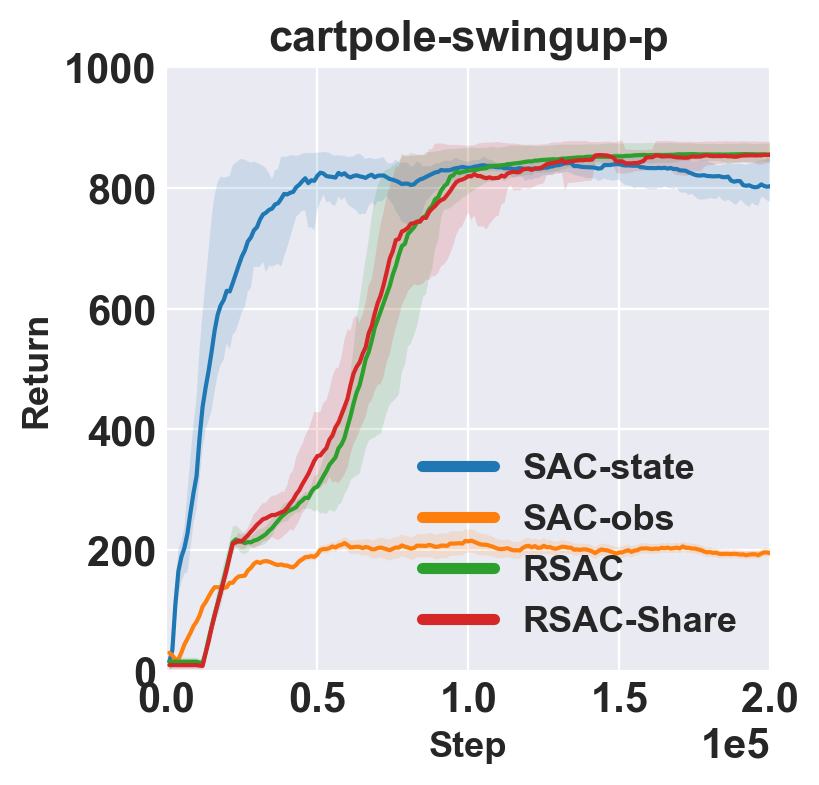}
        & \includegraphics[width=0.25\linewidth]{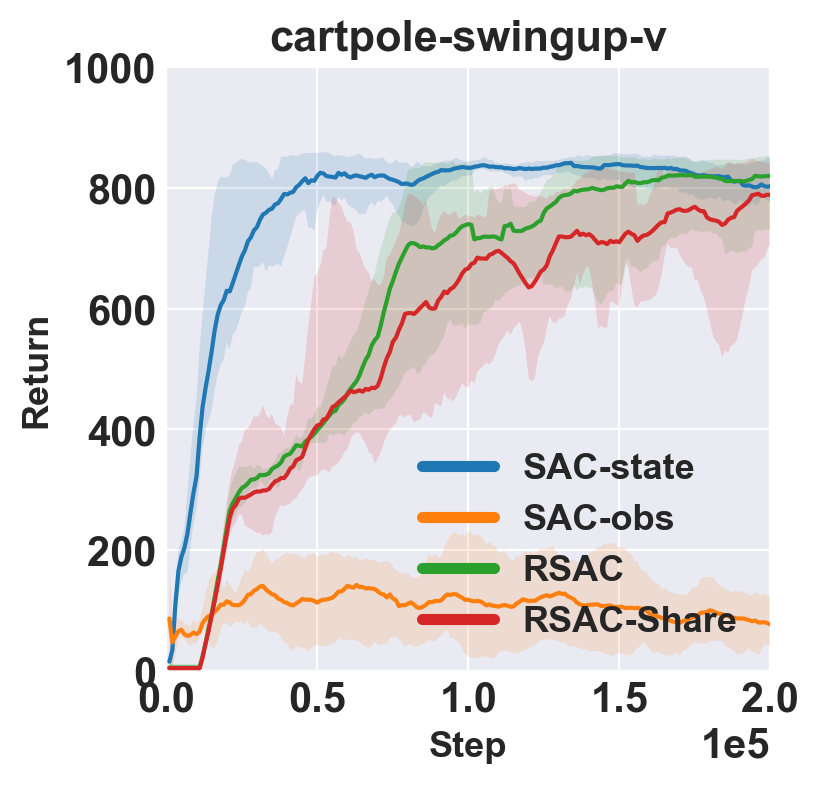}
        & \includegraphics[width=0.25\linewidth]{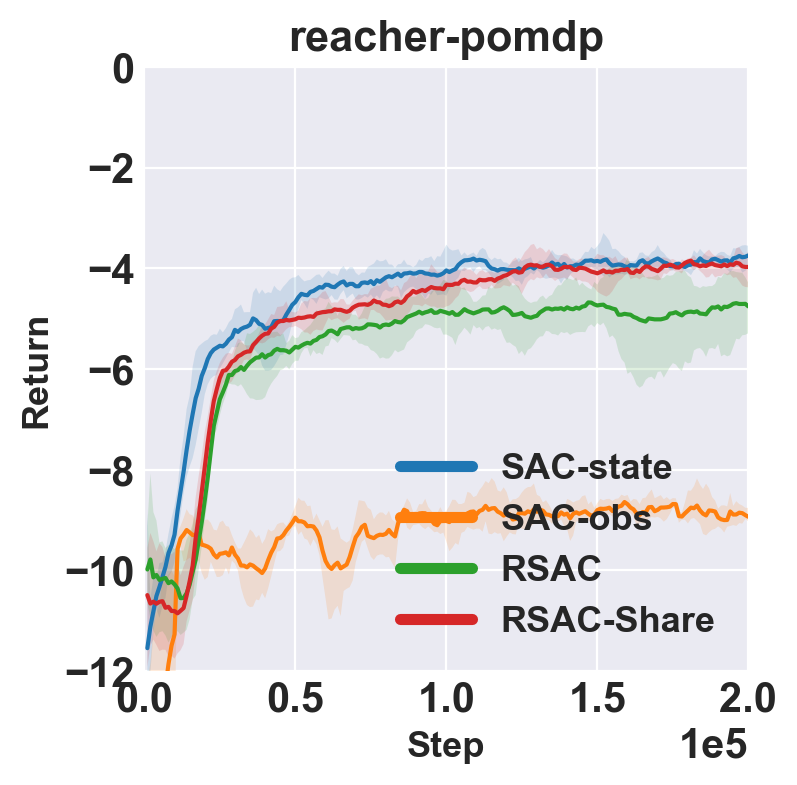}
        & \includegraphics[width=0.25\linewidth]{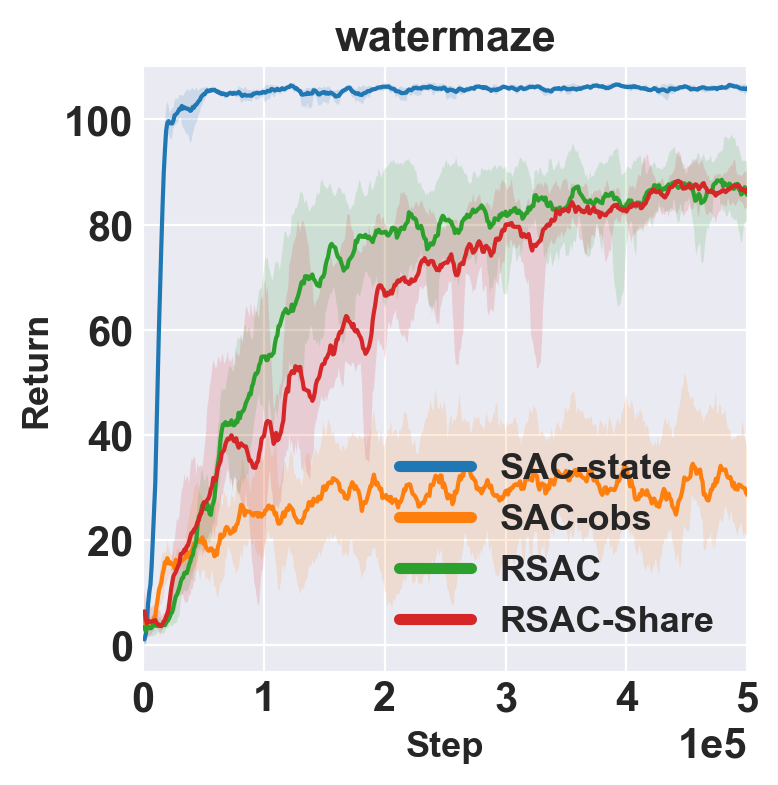}\\[-4pt]
        
        \includegraphics[width=0.25\linewidth]{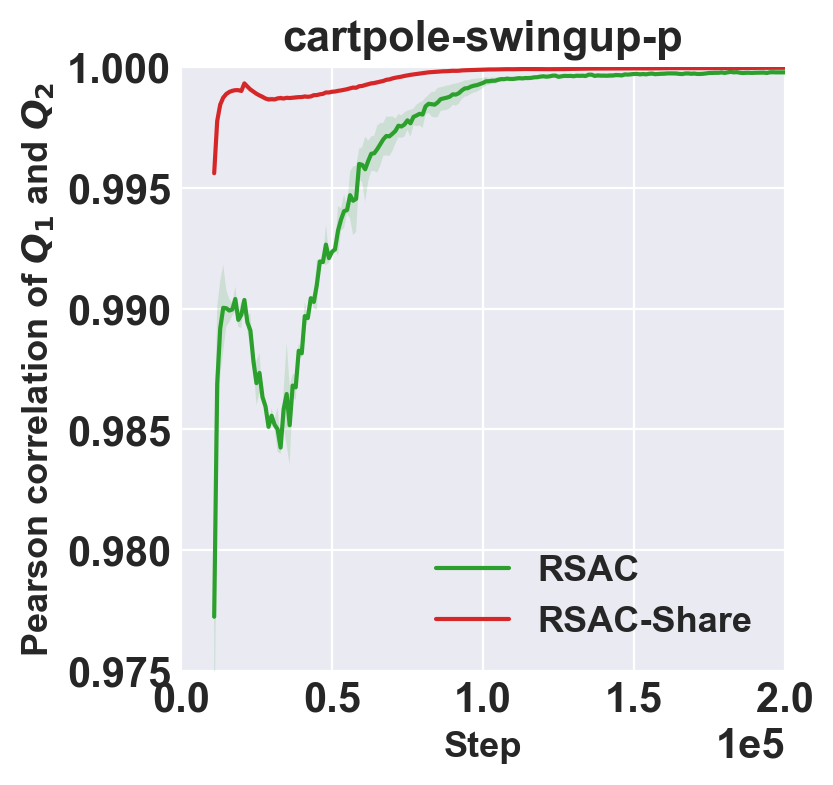}
        & \includegraphics[width=0.25\linewidth]{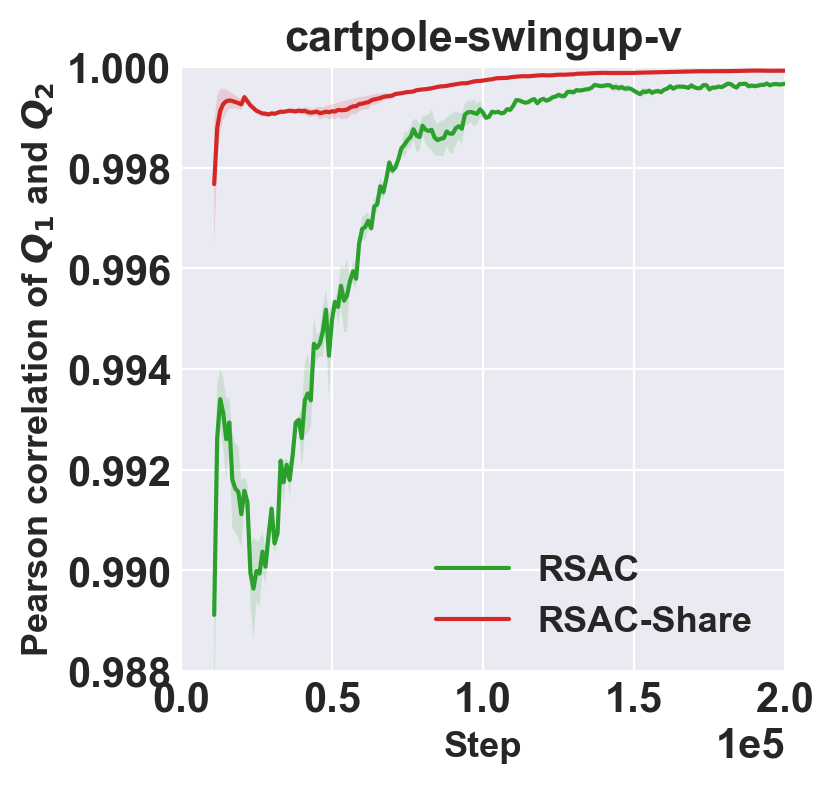}
        & \includegraphics[width=0.25\linewidth]{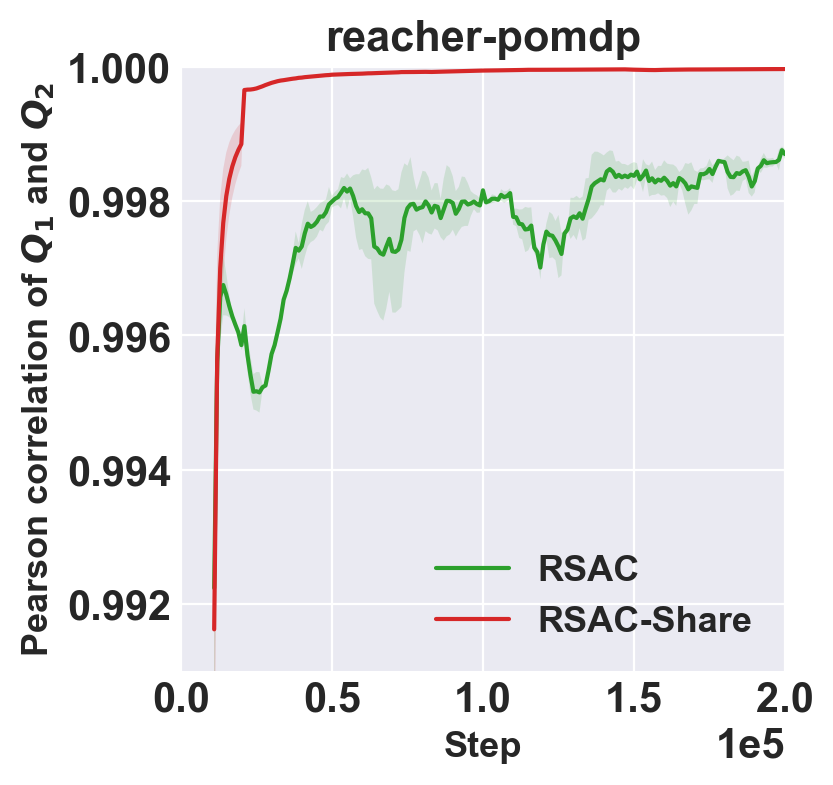}
        & \includegraphics[width=0.25\linewidth]{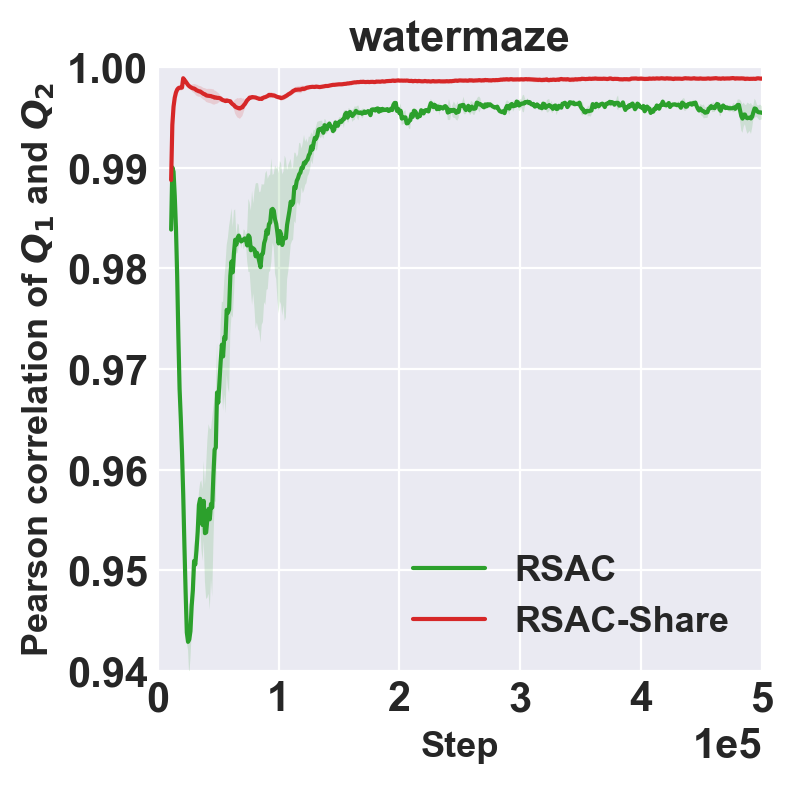}\\[-4pt]
        
        \includegraphics[width=0.25\linewidth]{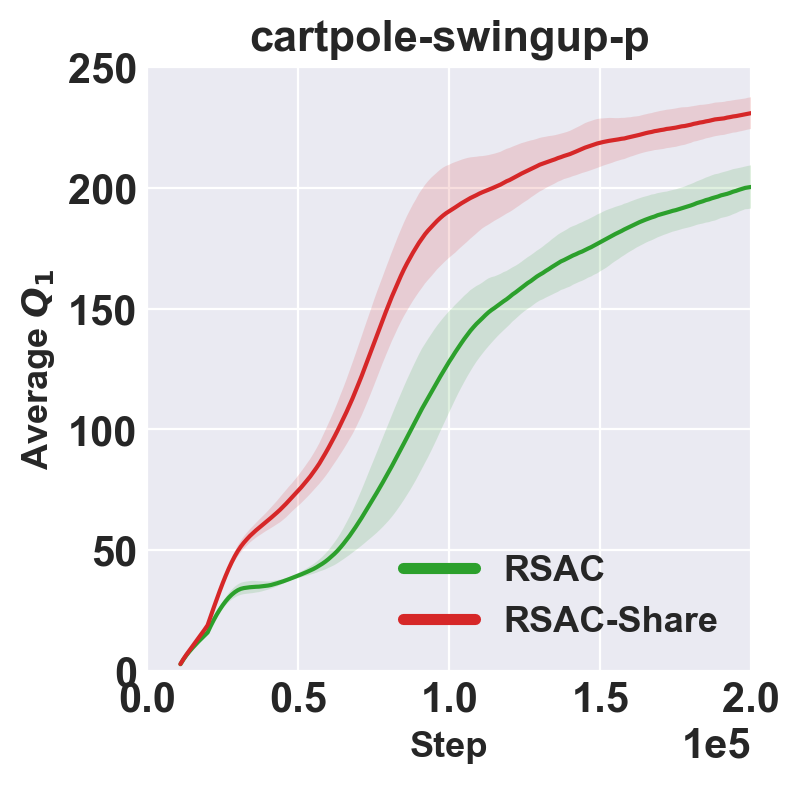}
        & \includegraphics[width=0.25\linewidth]{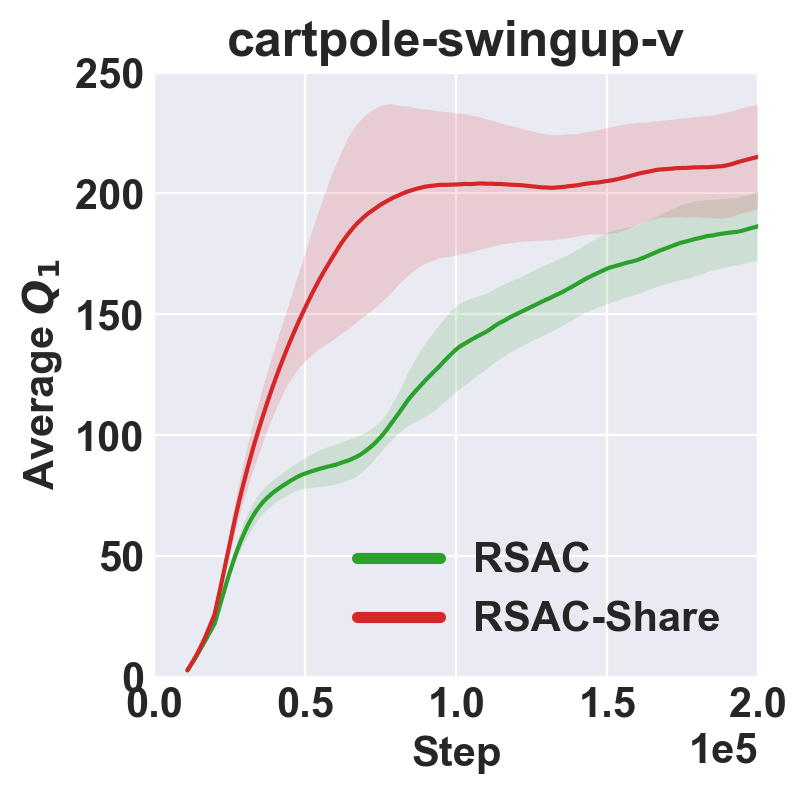}
        & \includegraphics[width=0.25\linewidth]{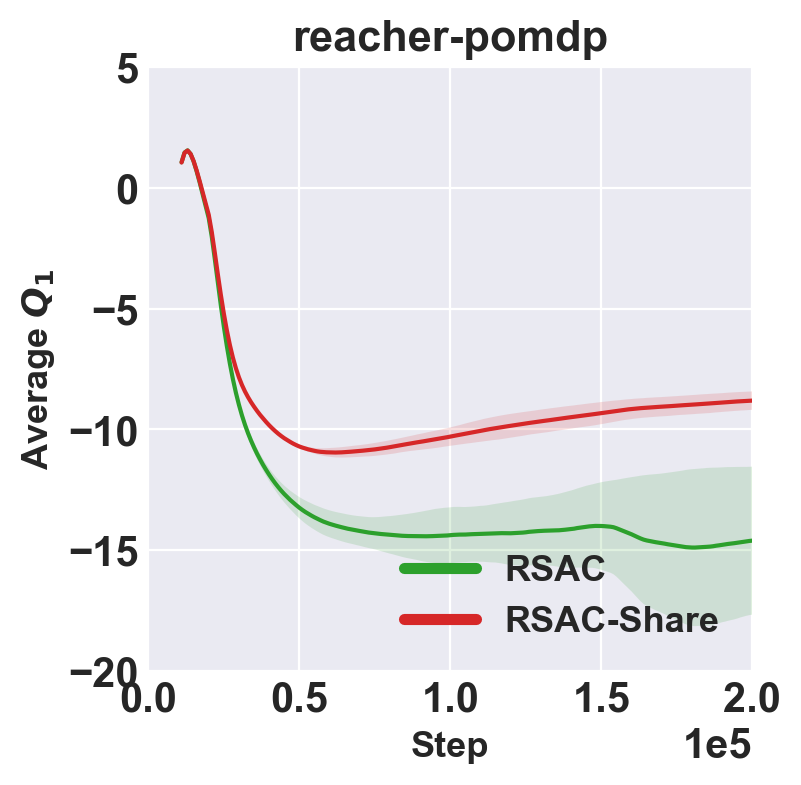}
        & \includegraphics[width=0.25\linewidth]{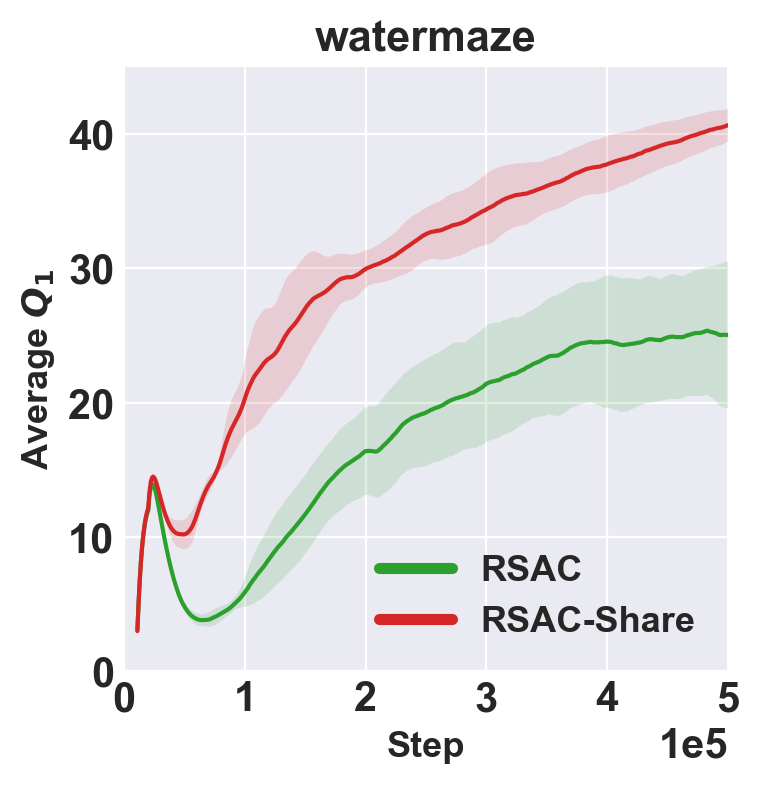}
    \end{tabular}%
    \caption{Performance of RSAC-Share against RSAC. The first row shows that RSAC-Share is on par with RSAC in terms of return. The second row shows that RSAC-Share has higher correlation between critics' outputs. The third row shows that RSAC-Share learn higher values. }
    \label{table:share}
    \vspace{-17pt}
\end{table}

\subsection{Comparison among VRNN, LSTM, and GRU} \label{section-compare}

In earlier experiments, we used LSTM as the recurrent representation since it has shown to be hard to beat on a variety of tasks \cite{greff2016lstm}, and was used as the recurrent component of several deep RL algorithms \cite{hausknecht2015deep, heess2015memory, han2019variational, meng2021memory}. In this section, we compare RSAC-LSTM, RSAC-VRNN, and RSAC-GRU against each other. Our experiments (see Table \ref{table:rnn_lstm_gru}) show that RSAC-LSTM significantly out-performs RSAC-VRNN and is on par with RSAC-GRU, indicating that using either LSTM or GRU as the recurrent representation is an equally reasonable choice. Interestingly, RSAC-GRU and RSAC-LSTM have different speeds of learning, but seem to achieve similar asymptotic performance. RSAC-VRNN performed the worst, presumably due to the well-known vanishing/exploding-gradient problem \cite{hochreiter2001gradient} when training VRNN using long sequences. 

\begin{table}
    \centering
    \setlength{\tabcolsep}{0pt}
    \begin{tabular}{cccc}
        \includegraphics[width=0.25\linewidth]{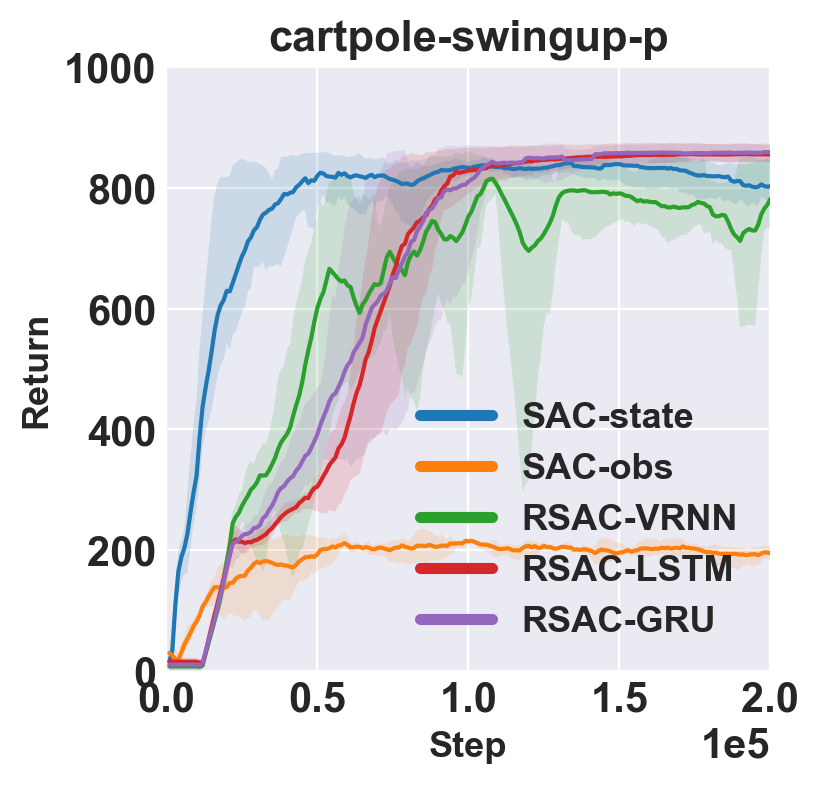}
        & \includegraphics[width=0.25\linewidth]{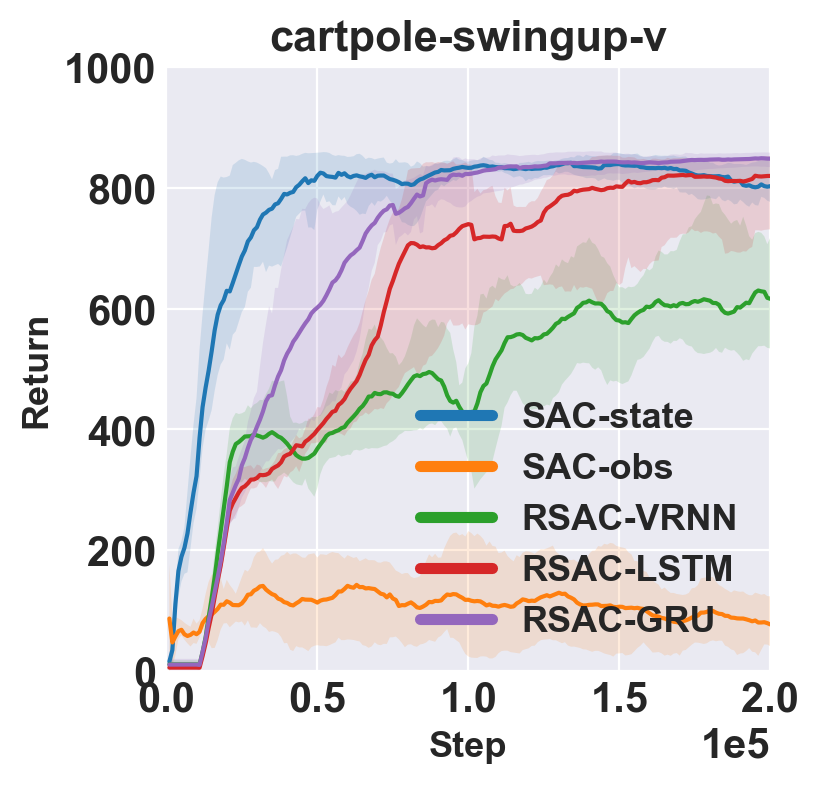}
        & \includegraphics[width=0.25\linewidth]{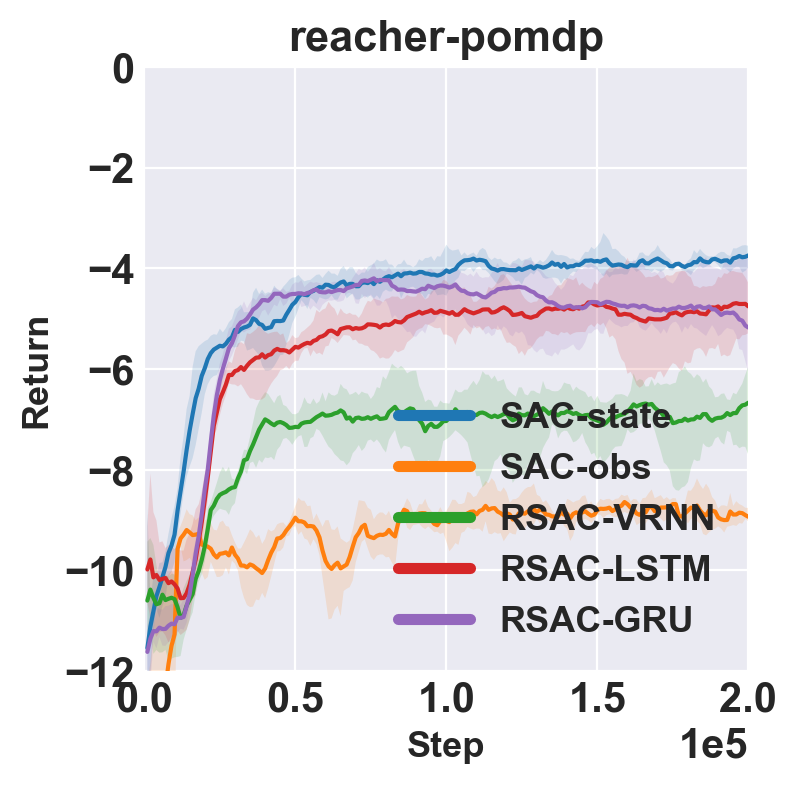}
        & \includegraphics[width=0.25\linewidth]{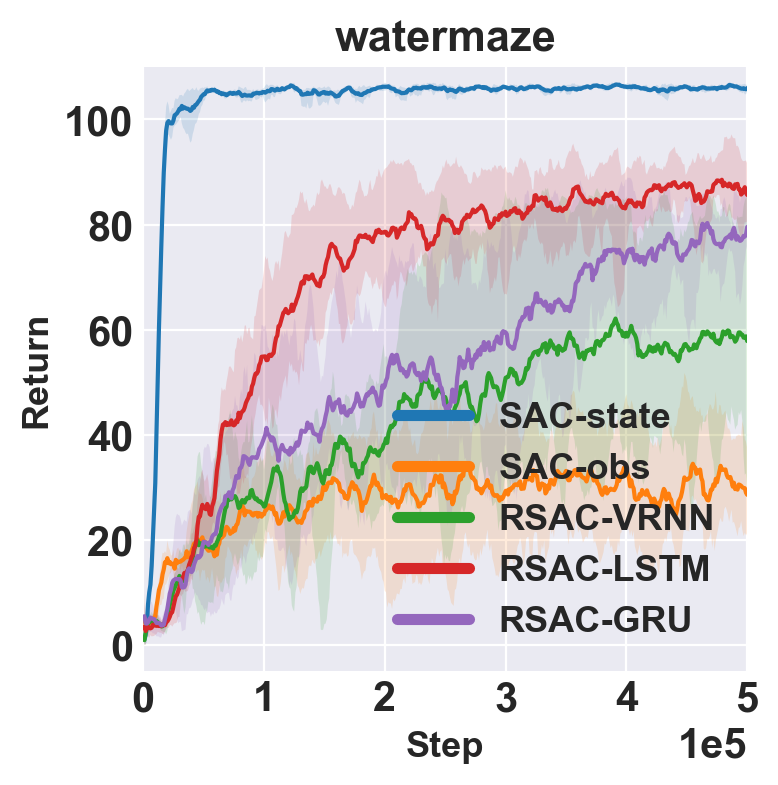}
    \end{tabular}%
    \caption{Performance of RSAC-LSTM, RSAC-VRNN, and RSAC-GRU. RSAC-LSTM and RSAC-GRU are on par with each other, and both greatly out-performs RSAC-VRNN.}
    \label{table:rnn_lstm_gru}
    \vspace{-17pt}
\end{table}

\section{Conclusions}\label{conclusion}


We present several recurrent off-policy baselines, tested on a diverse set of continuous control tasks. Our experiments show that RSAC is the most reliable algorithm out of the three baselines investigated RSAC, RDPG, and RTD3. However, they still face difficulties when solving tasks which provide only sparse rewards and require active exploration. Unfortunately, these two characteristics are often common in interesting real-world POMDPs. Another limitation is the inability to deal with too long episodes, inheriting from known weakness of RNNs (e.g., long training time, difficulties learning long-term dependencies). This suggests a potential research direction of using transformers \cite{vaswani2017attention}, which improve upon these weaknesses, as an alternative to RNNs for solving POMDPs.

\clearpage

\bibliographystyle{plainnat}
\bibliography{references.bib}

\clearpage

\clearpage

\appendix

\section{Appendix}



\subsection{Popular instantiations of RNNs}\label{appendix-rnn}

\paragraph{Elman Network (EN)} EN \cite{elman1990finding}, also known as the Vanilla RNN (VRNN), is one of the simplest forms of RNN. EN uses a single vector $h$ to represent the summary (of all previous inputs) and the output, and the parametrization used is
\begin{align*}
h_t = \tanh (W_{ih}x_t + b_{ih} + W_{hh}h_{t-1} + b_{hh}),
\end{align*}
where $W$ and $b$ denote learnable weights and biases. EN suffers from the vanishing-gradient problem \cite{hochreiter2001gradient}, which hinders its ability to learn long-term dependencies.

\paragraph{Long Short-term Memory (LSTM)} LSTM \cite{hochreiter1997long} was invented to tackle the vanishing-gradient problem that plagues EN. Even it seems to be handcrafted and is complicated, it has been shown to be hard to beat on a variety of tasks \cite{greff2016lstm}. Unlike EN, it uses two vectors $h$ (the hidden state) and $c$ (the cell state) to represent the summary (of all previous inputs), and only $h$ to represent the output; its parametrization is
\begin{align*} 
f_{t} &=\sigma\left(W_{i f} x_{t}+b_{i f}+W_{h f} h_{t-1}+b_{h f}\right) \tag{forget gate} \\
i_{t} &=\sigma\left(W_{i i} x_{t}+b_{i i}+W_{h i} h_{t-1}+b_{h i}\right) \tag{input gate} \\ 
\tilde{c}_{t} &=\tanh \left(W_{i g} x_{t}+b_{i g}+W_{h g} h_{t-1}+b_{h g}\right) \tag{candidate cell state} \\ 
c_{t} &=f_{t} \odot c_{t-1}+i_{t} \odot \tilde{c}_{t} \tag{new cell state} \\
o_{t} &=\sigma\left(W_{i o} x_{t}+b_{i o}+W_{h o} h_{t-1}+b_{h o}\right) \tag{output gate} \\
h_{t} &=o_{t} \odot \tanh \left(c_{t}\right), \tag{output} 
\end{align*}
where $W$ and $b$ denote learnable weights and biases. The parametrization has an intuitive appeal: the new cell state is the sum of the results of applying the forget gate to the new cell state and applying the input gate to the input.

\paragraph{Gated-recurrent Unit (GRU)} GRU \cite{cho2014learning} was motivated by LSTM but is simpler. Like EN, it uses a single vector $h$ (the hidden state) to represent the summary (of all previous inputs) and the output. Like LSTM, it computes the new hidden state by a weighted sum of the old hidden state and a candidate hidden state. The key difference is two-fold: how the candidate is computed, and the fact that GRU uses a single gate to replace two separate input and forget gates. GRU's parametrization is
\begin{align*} 
z_{t} &=\sigma\left(W_{i z} x_{t}+b_{i z}+W_{h z} h_{t-1}+b_{h z}\right) \tag{forget/input gate} \\
r_{t} &=\sigma\left(W_{i r} x_{t}+b_{i r}+W_{h r} h_{t-1}+b_{h r}\right) \\ 
\tilde{h}_{t} &=\tanh \left(W_{i g} x_{t}+b_{i g}+r_{t} *\left(W_{h g} h_{t-1}+b_{h g}\right)\right) \tag{candidate hidden state} \\ 
h_{t} &=\left(1-z_{t}\right) \tilde{h}_{t}+z_{t} * h_{t-1} \tag{hidden state \& output},
\end{align*}
where $W$ and $b$ denote learnable weights and biases.

\subsection{Comparison to VRM and its baselines}\label{rsac-vrm}

In this section, we compare our RSAC agent to VRM \cite{han2019variational} and its baselines: SAC-LSTM and SLAC. SAC-LSTM is similar to RSAC, but uses several tricks to allow for truncated BPTT so that it is more amendable to CPU training. While these tricks are useful, they have not been investigated thoroughly in literature. The version of SLAC used is a modification of the original SLAC algorithm \cite{lee2019stochastic} to non-pixel observations. On a set of four low-dimensional domains open-sourced by VRM authors, RSAC out-performs or is on par with all other algorithms (see Table \ref{table:rsac-vrm}). To promote fair comparison, we used the original code without any modifications, and plot the mean and standard error over 5 seeds as in the original paper.


\begin{table}[H]
    \centering
    \setlength{\tabcolsep}{0pt}
    \begin{tabular}{cccc}
        \includegraphics[width=0.25\textwidth]{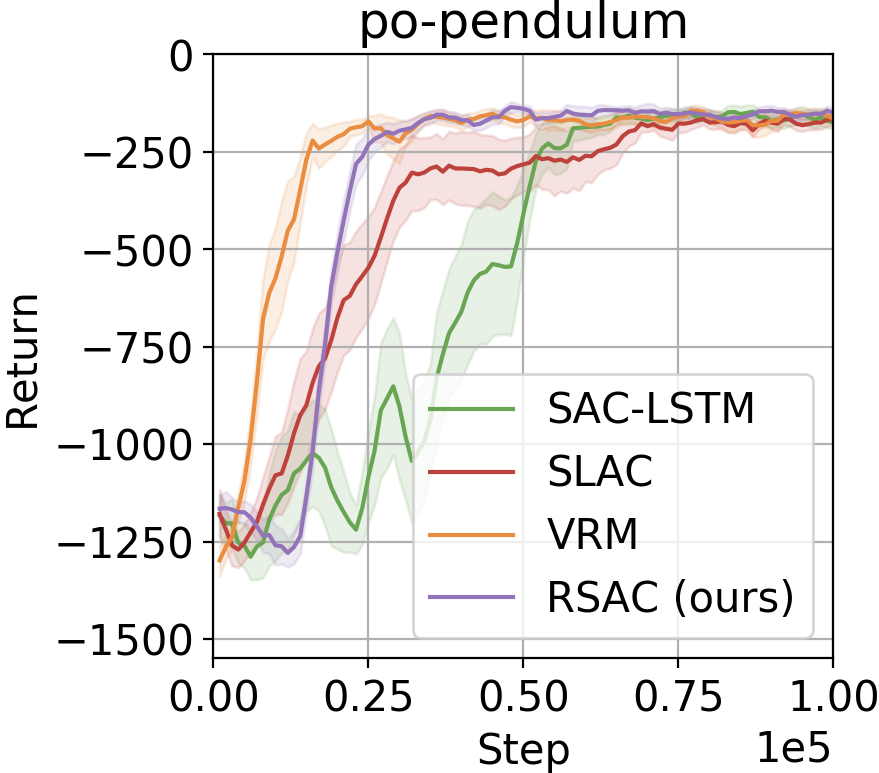}
        & \includegraphics[width=0.25\textwidth]{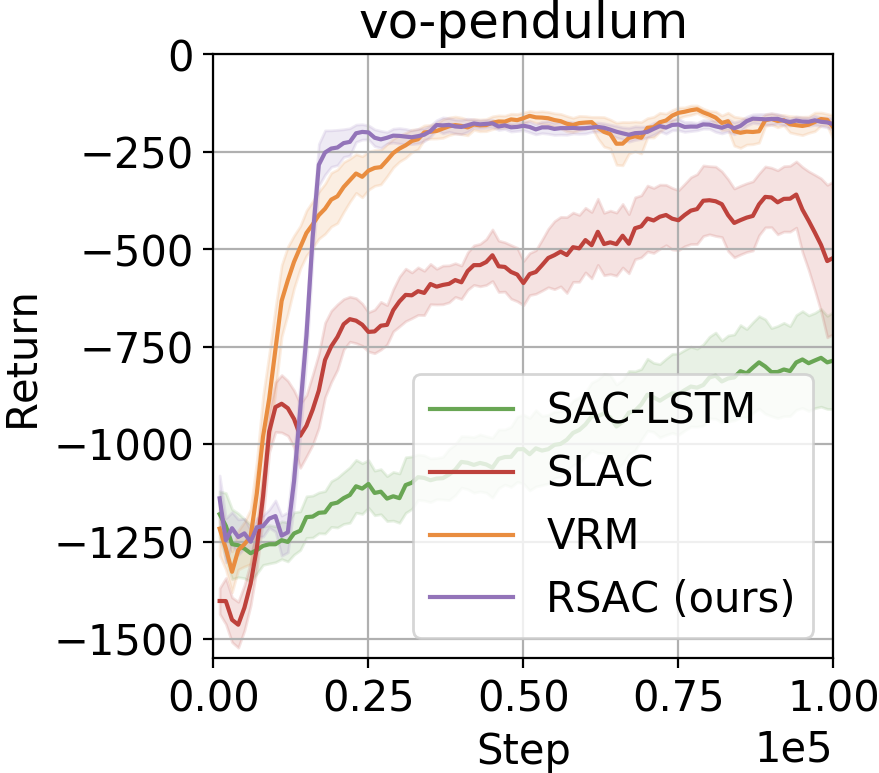}
        & \includegraphics[width=0.23\textwidth]{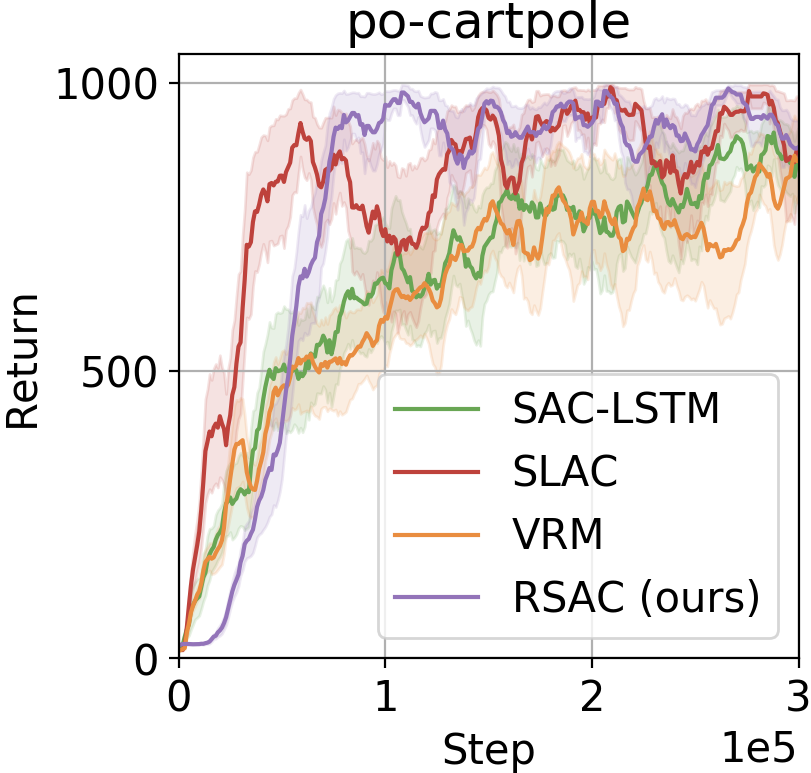}
        & \includegraphics[width=0.23\textwidth]{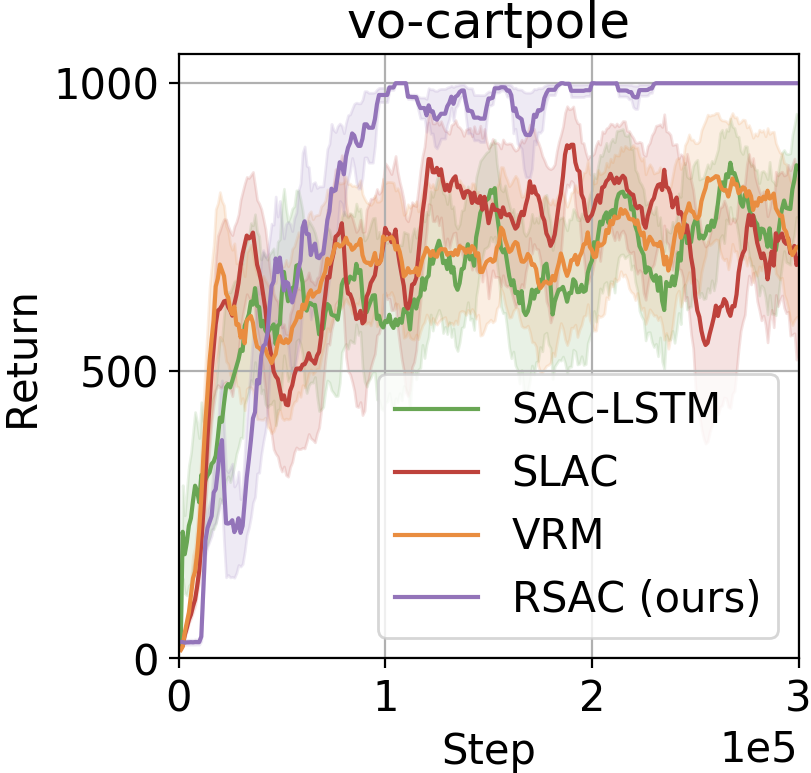}
    \end{tabular}%
    \caption{Performance of RSAC against VRM and its baselines.}
    \label{table:rsac-vrm}
    \vspace{-10pt}
\end{table}

\subsection{Hyper-parameters}\label{appendix-hyperparameter}

All our agents used hyper-parameters similar to a popular repository \cite{Raffin_Stable_Baselines3_2020}. 

For non-recurrent agents, the actors and critics are 2-layer perceptrons with hidden dimensions $(256, 256)$ with ReLU as the intermediate activation. For recurrent agents, two recurrent layers with hidden dimension $256$ are prepended to the non-recurrent actors and critics; note that we consider the recurrent layers as parts of the actors and critics.

For DDPG, TD3, and their recurrent versions, the action noise during training is $0.1$. For TD3 and its recurrent version, the target noise is $0.5$ and the noise clip is $0.2$. Note that these values are w.r.t. to a normalized action space such that each dimension lies within $[-1, 1]$. For SAC and its recurrent version, the entropy regularizer $\alpha$ is initialized to $1$ and auto-tuned as in \cite{Raffin_Stable_Baselines3_2020}, which implements \cite{haarnoja2018soft2}.

For non-recurrent agents, the replay buffer has a capacity of 1 million transitions and, for each network update, a batch of 100 transitions is sampled. For recurrent agents, the replay buffer has a capacity of 5000 episodes and, for each network update, a batch of 10 episodes is sampled. 
The discount factor $\gamma$ is $0.99$. The polyak-averaging $\rho$ coefficient is $0.995$. The learning rate for both actors and critics is $3e-4$. 

Non-recurrent algorithms take actions uniformly sampled from the action space for the first 1k steps, and recurrent algorithms do the same for the first 10k steps. During this period, returns are recorded but other statistics are unavailable, so we do not plot them.

\subsection{Pearson correlation coefficient}\label{pearson}

The correlation coefficient \cite{chihara2018mathematical}, or the Pearson correlation coefficient, is a popular measure for the strength of a \textit{linear} relationship between two random variables. If only pairs of samples $\mathcal{D}=\left\{(x_i, y_i)\right\}_{i=1}^M$ are available, we can compute the sample-version of this measure by the formula
\begin{align*}
    r_{\mathcal{D}} = \frac{\sum_{i=1}^M (x_i - \bar{x}) (y_i - \bar{y})}{\sqrt{\sum_{i=1}^M (x_i - \bar{x})^2} \sqrt{\sum_{i=1}^M (y_i - \bar{y})^2}},
\end{align*}
where $\bar{x}$ and $\bar{y}$ are the empirical means. 

\begin{figure}[H]
    \centering
    \includegraphics[width=1.0\textwidth]{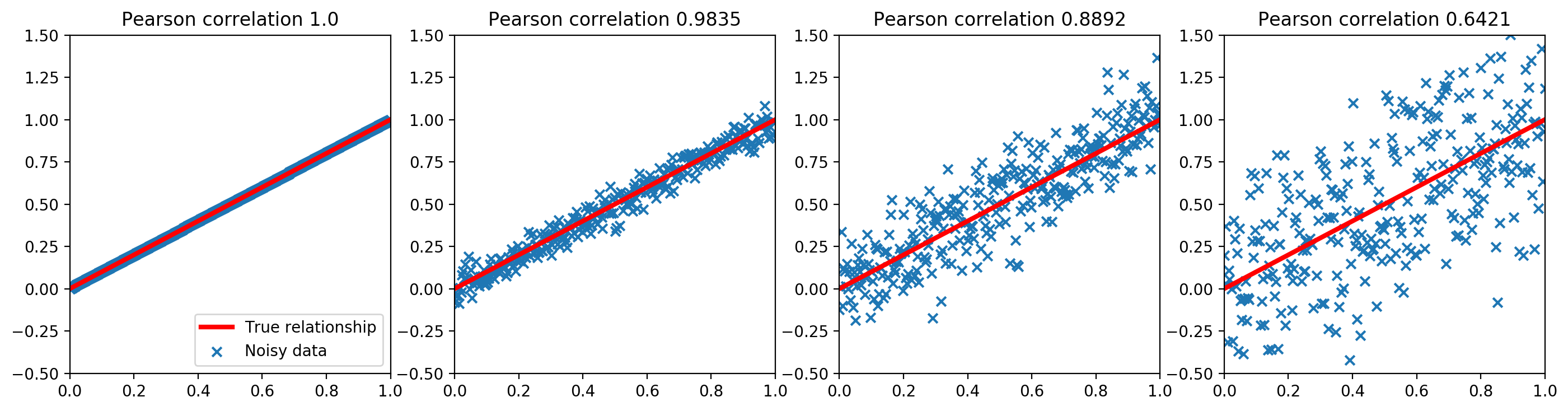}
    \caption{Pearson correlation for different example datasets.}%
    \label{fig:pearson}%
    \vspace{-10pt}
\end{figure}

We illustrate the result of applying this formula to four example datasets with decreasing relationship strength (see Figure \ref{fig:pearson}). In this work, we used $r$ under the assumption that $Q$-values produced by the two critics follow a linear relationship. We believe that this assumption is reasonable since both critics are updated using the same target, and hence have no reason to follow any other relationship.

\subsection{Additional trajectories for \texttt{push-r-bump}} \label{appendix-traj}

\begin{figure}[H]
    \centering
    \includegraphics[width=0.75\textwidth]{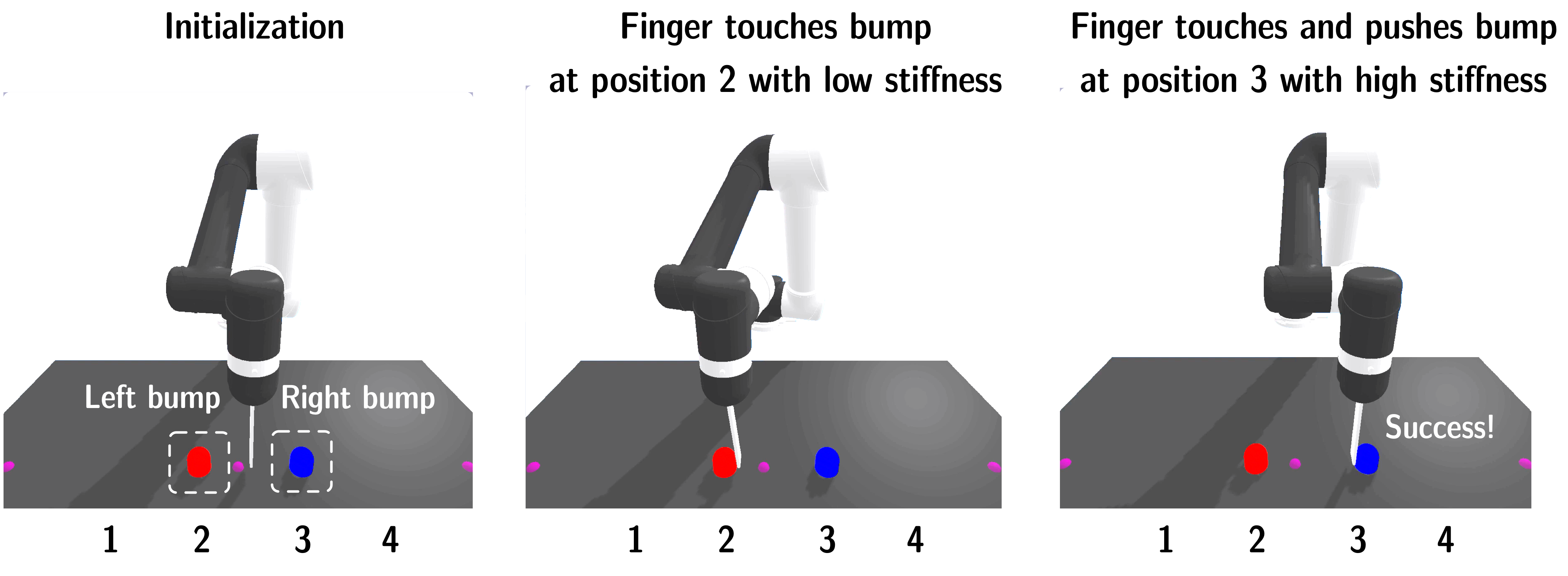}
    \caption{One RTD3 seed on Case 2 of \texttt{push-r-bump}.}%
    \label{fig:traj2}%
    \vspace{-10pt}
\end{figure}

\begin{figure}[H]
    \centering
    \includegraphics[width=1.0\textwidth]{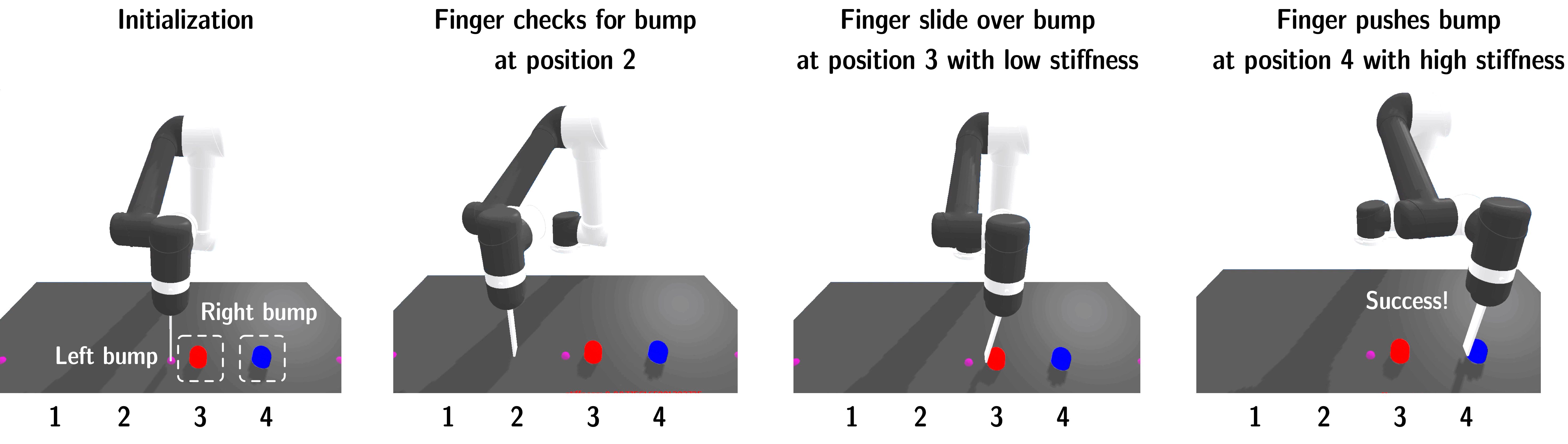}
    \caption{One RTD3 seed on Case 3 of \texttt{push-r-bump}.}%
    \label{fig:traj3}%
    \vspace{-10pt}
\end{figure}

In this section, we show trajectories of the same RTD3 seed used in Figure \ref{fig:policy_viz}c on two other cases of \texttt{push-r-bump}; the trajectories for Case 2 and 3 are shown in Figure \ref{fig:traj2} and \ref{fig:traj3} respectively. In all 3 cases, we observe that the agent first checks for bump at position 2, and then checks for bump at position 3. Intuitively, this behavior allows the agent to collect enough information to differentiate among the 3 cases: Case 1 has a bump at position 2 but not 3, Case 2 has a bump at both position 2 and 3, and Case 3 has a bump at position 3 but not 2. 

\begin{figure}[H]
    \centering
    \includegraphics[width=0.40\textwidth]{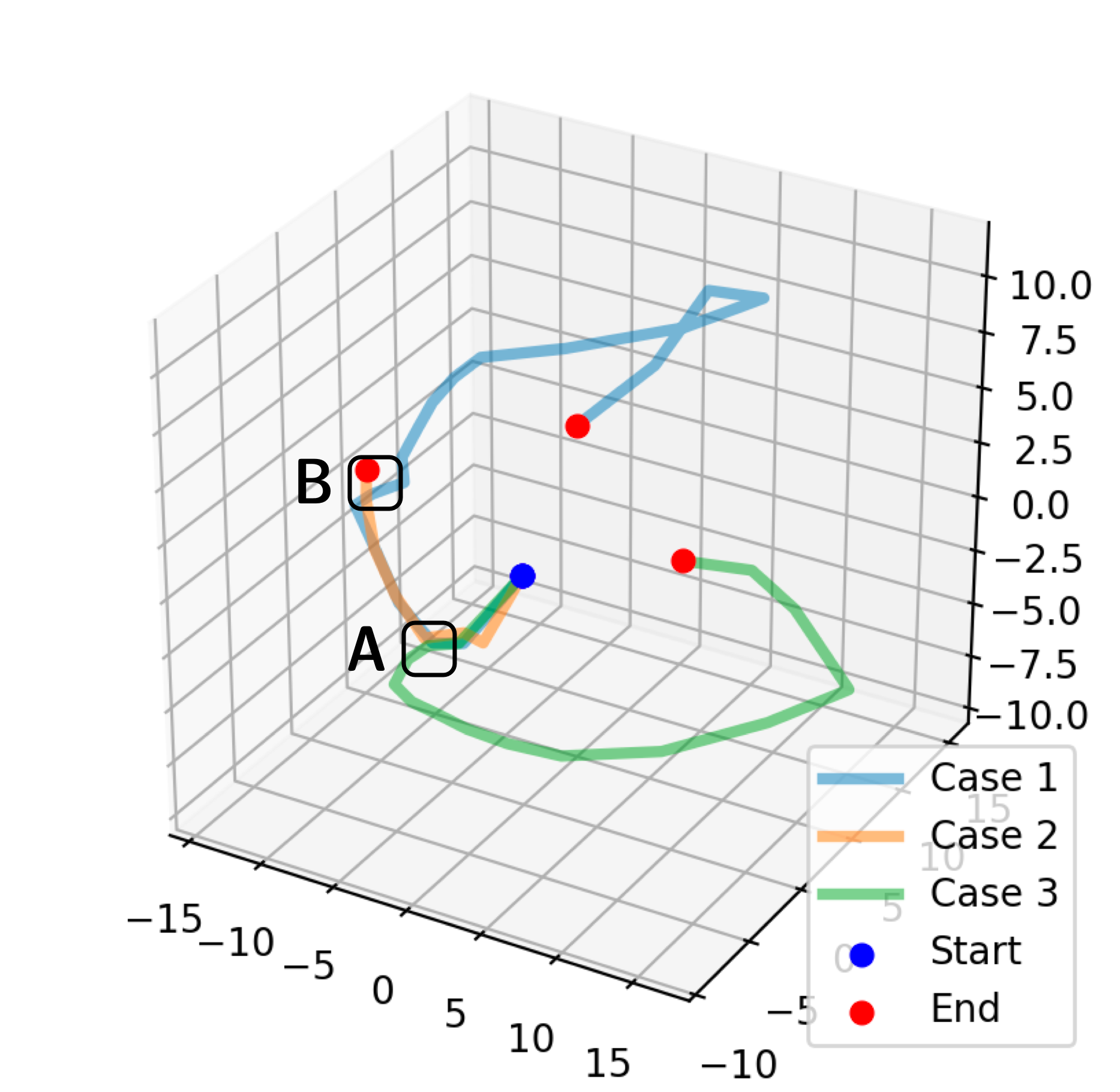}
    \caption{Evolution of the cell state of the 2nd LSTM layer (projected using PCA) of one RTD3 seed for all 3 cases in \texttt{push-r-bump}. Event A and B correspond to events when the finger makes contact with bumps. Axis scales are arbitrary.}%
    \label{fig:latent-vecs}%
    \vspace{-10pt}
\end{figure}

We also visualize the evolution of cell state of the 2nd LSTM layer of the same RTD3 seed for all 3 cases (see Figure \ref{fig:latent-vecs}). Cell states have 256 dimensions and we chose to project them to 3 dimensions using Principal Component Analysis (PCA). In this latent space, all trajectories share the same starting location, since the cell state is initialized to zeros. Case 3 splits with Case 1 and 2 at region A because the agent does not detect a bump at position 2. In region B, Case 2 terminates earlier than Case 1 because, in Case 2, the agent can directly push the bump at position 3 while, in Case 1, the agent does not detect a bump at position 3 and must return and push the bump at position 2.

\end{document}


\maketitle

\section{Appendix}

\subsection{Training and evaluation}

During training, the algorithms were evaluated for 10 episodes per 1000 steps of environment interactions and 1000 steps of network updates. The number of environment interactions and the number of network updates was locked to a 1-to-1 ratio, and the two events took place alternatively. During evaluation, we turned off stochasticity of actors.

\subsection{Hyper-parameters}

All our agents used hyper-parameters similar to a popular repository \cite{Raffin_Stable_Baselines3_2020}. 

For non-recurrent agents, the actors and critics were multi-layer perceptrons with hidden dimensions $(256, 256)$ with ReLU as the intermediate activation. For recurrent agents, two recurrent layers with hidden dimension $256$ were prepended to the non-recurrent actors and critics; note that we consider the recurrent layers as parts of the actors and critics.

For DDPG, TD3 and their recurrent versions, the action noise during training was $0.1$. For TD3 and its recurrent version, the target noise was $0.5$ and the noise clip was $0.2$. Note that these values are w.r.t. to a normalized action space such that each dimension lies within $[-1, 1]$. For SAC and its recurrent version, the entropy regularizer $\alpha$ was initialized to $1$ and auto-tuned as \cite{Raffin_Stable_Baselines3_2020}.

For non-recurrent agents, the replay buffer had a capacity of 1 million transitions and, for each network update, a batch of 100 transitions was sampled. For recurrent agents, the replay buffer had a capacity of 5000 episodes and, for each network update, a batch of 10 episodes was sampled. 

The discount factor $\gamma$ was $0.99$. The polyak-averaging $\rho$ coefficient was $0.995$. The learning rate for both actors and critics was $3e-4$. 

\subsection{Popular instantiations of RNNs}

\paragraph{Elman Network (EN)} EN \cite{elman1990finding}, also known as the Vanilla RNN (VRNN), is one of the simplest forms of RNN. EN uses a single vector $h$ to represent the summary and the output, and the parametrization used is \vspace{2mm}
$$h_t = \tanh (W_{ih}x_t + b_{ih} + W_{hh}h_{t-1} + b_{hh}),$$

where $W$ and $b$ denote learnable weights and biases. EN suffers from the vanishing-gradient problem \cite{hochreiter2001gradient}, which hinders its ability to learn long-term dependencies.

\paragraph{Long Short-term Memory (LSTM)} LSTM \cite{hochreiter1997long} was invented to tackle the vanishing-gradient problem that plagues EN. Despite it seems to be handcrafted and is complicated, it has been shown to be hard to beat on a variety of tasks \cite{greff2016lstm}. Unlike EN, it uses two vectors $h$ (the hidden state) and $c$ (the cell state) to represent the summary, and only $h$ to represent the output; its parametrization is \vspace{2mm}
\begin{align*} 
f_{t} &=\sigma\left(W_{i f} x_{t}+b_{i f}+W_{h f} h_{t-1}+b_{h f}\right) \tag{forget gate} \\
i_{t} &=\sigma\left(W_{i i} x_{t}+b_{i i}+W_{h i} h_{t-1}+b_{h i}\right) \tag{input gate} \\ 
\tilde{c}_{t} &=\tanh \left(W_{i g} x_{t}+b_{i g}+W_{h g} h_{t-1}+b_{h g}\right) \tag{candidate cell state} \\ 
c_{t} &=f_{t} \odot c_{t-1}+i_{t} \odot \tilde{c}_{t} \tag{new cell state} \\
o_{t} &=\sigma\left(W_{i o} x_{t}+b_{i o}+W_{h o} h_{t-1}+b_{h o}\right) \tag{output gate} \\
h_{t} &=o_{t} \odot \tanh \left(c_{t}\right), \tag{output} 
\end{align*}

where $W$ and $b$ denote learnable weights and biases. The parametrization has an intuitive appeal: the new cell state is the sum of the results of applying the forget gate to the new cell state and applying the input gate to the input.

\paragraph{Gated-recurrent Unit (GRU)} GRU \cite{cho2014learning} was motivated by LSTM but is simpler. Like EN, it uses a single vector $h$ (the hidden state) to represent the summary and the output. Like LSTM, it computes the new hidden state by a weighted sum of the old hidden state and a candidate hidden state. The key difference is two-fold: how the candidate is computed, and the fact that GRU uses a single gate to replace two separate input and forget gates. GRU's parametrization is \vspace{2mm}
\begin{align*} 
z_{t} &=\sigma\left(W_{i z} x_{t}+b_{i z}+W_{h z} h_{t-1}+b_{h z}\right) \tag{forget/input gate} \\
r_{t} &=\sigma\left(W_{i r} x_{t}+b_{i r}+W_{h r} h_{t-1}+b_{h r}\right) \\ 
\tilde{h}_{t} &=\tanh \left(W_{i g} x_{t}+b_{i g}+r_{t} *\left(W_{h g} h_{t-1}+b_{h g}\right)\right) \tag{candidate hidden state} \\ 
h_{t} &=\left(1-z_{t}\right) \tilde{h}_{t}+z_{t} * h_{t-1} \tag{hidden state \& output},
\end{align*}

where $W$ and $b$ denote learnable weights and biases.

\bibliographystyle{plainnat}
\bibliography{references.bib}